\pdfoutput=1
\documentclass[11pt]{article}

\usepackage{emnlp2021}

\usepackage{times}
\usepackage{latexsym}
\newtheorem{theorem}{Theorem}[section]

\newtheorem{lemma}[theorem]{Lemma}
\usepackage[T1]{fontenc}
\usepackage[utf8]{inputenc}
\usepackage{graphicx}
\usepackage{todonotes}
\usepackage{amssymb}
\usepackage{amsmath}
\usepackage{mathtools}
\usepackage{booktabs}
\usepackage{caption}
\usepackage{subcaption}
\usepackage{microtype}
\usepackage{placeins}
\usepackage{balance}
\newcommand\numberthis{\addtocounter{equation}{1}\tag{\theequation}}

\DeclarePairedDelimiter{\floor}{\lfloor}{\rfloor}

\newcommand{\RNum}[1]{\uppercase\expandafter{\romannumeral #1\relax}}
\newcommand{\Rnum}[1]{\expandafter{\romannumeral #1\relax}}
\usepackage[ruled,linesnumbered]{algorithm2e}

\newcommand*{\affaddr}[1]{#1} % No op here. Customize it for different styles.
\newcommand*{\affmark}[1][*]{\textsuperscript{#1}}
\newcommand*{\email}[1]{\texttt{#1}}

\title{Distiller: A Systematic Study of Model Distillation Methods in Natural Language Processing}

\author{
Haoyu He\affmark[1]\thanks{\ \ Work done while being an intern at Amazon Web Services.}, Xingjian Shi\affmark[2], Jonas Mueller\affmark[2], Sheng Zha\affmark[2], Mu Li\affmark[2], and George Karypis\affmark[2]\\
\affaddr{\affmark[1]Northeastern University} \affaddr{\affmark[2]Amazon Web Services}\\
\email{he.haoy@northeastern.edu}\\
\email{\{xjshi,jonasmue,zhasheng,mli,gkarypis\}@amazon.com}\\
}
\begin{document}
\maketitle

\begin{abstract} % Old Abstract: Too long!
Knowledge Distillation (KD) offers a natural way to reduce the latency and memory/energy usage of massive  pretrained models that have come to dominate Natural Language Processing (NLP) in recent years.
While numerous sophisticated variants of KD algorithms have been proposed for NLP applications, the key factors underpinning the optimal distillation performance are often confounded and remain unclear.
We aim to identify how different components in the KD pipeline affect the resulting performance and how much the optimal KD pipeline varies across different datasets/tasks, such as the data augmentation policy, the loss function, and the intermediate representation for transferring the knowledge between teacher and student. %Here we conduct a systematic study of KD algorithms in NLP, aiming to identify how different components in the KD pipeline affect the resulting performance and how much the optimal KD pipeline varies across different datasets/tasks. 
% To systematically answer these questions, we propose \emph{Distiller}, a meta KD framework that covers a broad range of techniques across each stage of the KD pipeline. 
To tease apart their effects, we propose Distiller, a meta KD framework that systematically combines a broad range of  techniques across different stages of the KD pipeline, which enables us to quantify each component's contribution. 
Within Distiller, we unify commonly used objectives for distillation of intermediate representations under a universal mutual information (MI) objective 
% To ensure \emph{Distiller} is broadly applicable across different teacher/student architectures, we unify commonly used objectives for distillation of intermediate representations under a universal mutual information (MI) objective 
and propose a class of \emph{MI-$\alpha$} objective functions with better bias/variance trade-off for estimating the MI between the teacher and the student. 
  On a diverse set of NLP datasets, the best \emph{Distiller} configurations are identified via large-scale hyper-parameter optimization. 
 Our experiments reveal the following: 1) the approach used to distill the intermediate  representations is the most important factor in KD performance, 2) among different objectives for intermediate distillation, MI-$\alpha$ performs the best, and 3) data augmentation provides a large boost for small training datasets or small student networks. Moreover, we find that different datasets/tasks prefer different KD algorithms, and thus propose a simple \emph{AutoDistiller} algorithm that can recommend a good KD pipeline for a new dataset.
\end{abstract}

\section{Introduction}

Recent advancements in Natural Language Processing (NLP) such as  BERT~\cite{devlin2018bert}, RoBERTa~\cite{liu2019roberta}, and ELECTRA~\cite{clark2020electra} have demonstrated the effectiveness of Transformer models in generating transferable language representations. 
Pretraining over a massive unlabeled corpus and then fine-tuning over labeled data for the task of interest has become the state-of-the-art paradigm for solving diverse NLP problems ranging from sentence classification to question answering~\cite{raffel2019exploring}. 
Scaling up the size of these networks has led to rapid NLP improvements. However, these improvements have come at the expense of significant increase of memory for their many parameters and compute to produce predictions \cite{gpt3,kaplan2020scaling}.
% attributed to their large model size and high FLOPs, the promising results offered by these pre-trained models come with high computational cost and memory usage. 
This prevents these models from being deployed on resource-constrained devices such as smart phones and browsers, or for latency-constrained applications such as click-through-rate prediction. There is great demand for models that are smaller in size, yet still retain similar accuracy as those having a large number of parameters.%, which is critical for  application in broader scenarios.

To reduce the model size while preserving accuracy, various model compression techniques have been proposed such as:  pruning, quantization, and Knowledge Distillation (KD)~\cite{gupta2020compression}. Among these methods, \emph{task-aware} KD is a popular and particularly promising approach for compressing Transformer-based models~\cite{gupta2020compression}. The general idea is to first fine-tune a large model (namely the  \emph{teacher} model) based on the task's labeled data, and then train a separate network that has significantly fewer parameters (namely the   \emph{student} model) than the original model to mimic the predictions of the original large model.
% The original fine-tuned model is called \emph{teacher} and the smaller is called the \emph{student}, such that the student model can subsequently be deployed in place of the teacher.
% approach is to first obtain the teacher network by finetuning a large-scale pre-trained model on the downstream task and then distill the teacher’s knowledge to a light-weighted student network. 
A large number of task-aware KD algorithms have been proposed in the NLP regime, e.g.,  DistillBERT~\cite{sanh2019distilbert}, BERT-PKD~\cite{sun2019patient}, BERT-EMD~\cite{li2020bert}, TinyBERT~\cite{jiao2019tinybert}, DynaBERT~\cite{hou2020dynabert}, and AdaBERT~\cite{chen2020adabert}, some of which can compress the teacher network by 10$\times$ without significant accuracy loss on certain datasets.

Innovations in KD for NLP generally involve improvements in one of the following aspects: 1) the loss function for gauging the discrepancy between student and teacher predictions~\cite{kim2021comparing}, 2) the method for transferring intermediate network representations between teacher and student~\cite{sun2019patient,li2020bert,yang2020knowledge}, 3) the use of data augmentation during student training~\cite{jiao2019tinybert}, and 4) multiple stages of  distillation~\cite{chen2020adabert,mirzadeh2020improved}. 
Many research proposals have simultaneously introduced new variations of more than one of these components, which confounds the impact of each component on the final performance of the distillation algorithm. In addition, it is often unclear whether a proposed distillation pipeline will generalize to a new dataset or task, making automated KD challenging. 
For example, MixKD~\cite{liang2020mixkd} has only been evaluated on classification problems and it is unclear if the method will be effective for question answering.
%Moreover, important phenomena regarding KD for NLP remain unexplained, such as the observation by \citet{sun2019patient} that distilling from a larger teacher can make the student perform worse in some down-stream tasks. Much popular KD work~\cite{li2020bert,sanh2019distilbert} chose to distill from the BERT-base model and has not  investigated if their proposed methods still perform well for larger teachers. 

To understand the importance of different components in KD, we undertake a systematic study of KD algorithms in NLP. Our study is conducted using a meta-distillation pipeline we call  \emph{Distiller} that contains multiple configurable components. All candidate algorithms in the search space of \emph{Distiller} work for two types of NLP tasks: text classification and sentence tagging. 
Distiller unifies existing techniques for knowledge transfer from intermediate layers of the teacher network to the student network (i.e.\ \emph{intermediate distillation}) as special cases of maximizing (bounds on) the Mutual Information (MI) between teacher and student representations.
%We prove that existing objectives for intermediate distillation can be viewed as maximizing different lower bounds of MI. 
Based on this unification and recent progress in variational bounds of MI~\cite{poole2019variational}, we propose a new intermediate distillation objective called MI-$\alpha$ that uses a scalar $\alpha$ to control the bias-variance trade-off of MI estimation. Including MI-$\alpha$ in the search space of \emph{Distiller}, we run extensive hyper-parameter tuning algorithms to search for the best  \emph{Distiller}-configuration choices over  GLUE~\cite{wang2018glue} and SQuAD~\cite{rajpurkar2016squad}. 
This search helps us identify the best distillation pipelines and understand what impact different KD modules have on student performance in NLP. 
% Jonas: this is already mentioned in abstract. There are three major findings from our experiments: 1) among all components that we study, the design of the intermediate distillation module is the most important, 2) MI-$\alpha$ consistently outperforms the other objective functions, and 3) data augmentation gives significant boosts for small students and datasets. In addition, we find that the best configuration is often dataset/task specific.
%and the pipeline searched by \emph{Distiller} achieves state-of-the-art performance on GLUE and SQuAD.
%Thus, the data points acquired by our study can be used as a dataset to facilitate research on automated KD. To demonstrate such usage, 
Using the observations of this large-scale study, we train a \emph{AutoDistiller} model to predict the \emph{distillation ratio}, which is defined as the fraction of the teacher's performance achieved by the student, based on KD pipeline choices and characteristics of a dataset. Leave-one-out cross validation evaluation of \emph{AutoDistiller} demonstrates that it is able to reliably prioritize high-performing KD configurations in most folds, and is able to suggest good distillation pipelines on two new datasets. 

The main contributions of this work include:
\begin{itemize}
    \item The meta KD pipeline \emph{Distiller} used to systematically study the impact of different components in KD, including the: 1) data augmentation policy, 2) loss function for transferring intermediate representations, 3) layer mapping strategies for intermediate representations, 4) loss function for transferring outputs, as well as what role the task/dataset play.
    \item Unification of existing objectives for distilling intermediate representations as instances of maximizing bounds of the mutual information between teacher and student representations. 
    % A unifying perspective on existing objectives for distilling intermediate representations as all maximizing the mutual information between teacher and student representations. Based on this unification, we propose 
    This leads us to propose  
    the MI-$\alpha$ objective that outperforms the existing objectives.
    \item Using the results collected from our systematic \emph{Distiller} study, we fit a model that automatically predicts the best distillation strategy for a new dataset. On a never-seen dataset ``BoolQ''\cite{wang2019superglue},  predicted strategies achieve $1.002$ distillation ratios (fraction of the student's and the teacher's performance) on average, outperforming random selected strategies with mean of $0.960$. To the best of our knowledge, this is the first attempt towards automated KD in NLP.
\end{itemize}
\section{Related Work}
\label{related work}
\paragraph{Knowledge Distillation.}
The general KD framework was popularized by \citet{bucilua2006model,hinton2015distilling}, aiming to transfer knowledge from an accurate but cumbersome teacher model to a compact student model by matching the class probabilities produced by the teacher and the student. 
Focusing on AutoML settings with tabular data, \citet{fakoor2020fast} proposed a general KD algorithm for different classical ML models and ensembles thereof.
Also hoping to identify good choices in the KD pipeline like our work, \citet{kim2021comparing} compared Kullback-Leibler divergence and mean squared error objectives in KD for image classification models, finding that mean squared error performs better. 

Recent KD research in the domain of NLP has investigated how to efficiently transfer knowledge from pretrained Transformer models. \citet{sun2019patient} proposed BERT-PKD that transfers the knowledge from both the final layer and intermediate layers of the teacher network. \citet{jiao2019tinybert} proposed the TinyBERT model that first distills the general knowledge of the teacher by minimizing the Masked Language Model (MLM) objective~\cite{devlin2018bert}, with subsequent  task-specific distillation. \citet{li2020bert} proposed a many-to-many layer mapping function leveraging the Earth Mover’s Distance to transfer intermediate knowledge.
% \citet{hou2020dynabert} proposed DynaBERT, a multistage distillation pipeline to distill the teacher Transformer to student Transformers with different widths and depths. 
% \citet{mukherjee-hassan-awadallah-2020-xtremedistil} also considered multistage KD pipelines with a focus on multilingual Named Entity Recognition. 
% \citet{xu2021bert} proposed NAS-BERT which leverages neural architecture search to train a big super-network that can output multiple compressed models with adaptive sizes and latency. 
Our paper differs from the existing work in that we provide a systematic analysis of the different components in  single-stage task-aware KD algorithms in NLP and propose the first automated KD algorithm in this area.
% as well as the use of mutual information objectives for intermediate representation distillation that outperform the existing objectives.
\paragraph{Mutual Information Estimation.}
Mutual Information (MI) measures the degree of statistical dependence between random variables. Given random variables $A$ and $B$, the MI between them, $I(A, B)$, can be understood as how much knowing $A$ will reduce the uncertainty of $B$ or vice versa. For distributions that do not have analytical forms, maximizing MI directly is often intractable.
%when $A$ and $B$ are dense representations.
To overcome this difficulty, recent work resorts to variational bounds~\cite{donsker1975asymptotic, blei2017variational, nguyen2010estimating} and deep learning~\cite{oord2018representation} to estimate MI. These works utilize flexible parametric distributions or \emph{critics} that are parameterized neural networks~(NNs) to approximate unknown densities that appear in MI calcuations. \citet{poole2019variational} provides a review of existing MI estimators and proposes novel bounds that trade-off bias and variance. \citet{kong2019mutual} unified language representation learning objective functions from the MI maximization perspective.
% Our paper differs from this work in that we unify diverse objectives used for distilling intermediate representations, a generalized perspective that leads us to a better-performing MI-$\alpha$ objective.
\paragraph{Data Augmentation in NLP.}
Data Augmentation (DA) is an effective technique for improving the accuracy of text classification models~\cite{wei2019eda} and has also been shown to boost the performance of KD for NLP algorithms~\cite{jiao2019tinybert} as well as KD in models for tabular data~\cite{fakoor2020fast}. \citet{wei2019eda}  proposed the Easy Data Augmentation (EDA) technique that randomly replaces synonyms, inserts, swaps and deletes characters in the sentence. \citet{jiao2019tinybert} proposed to utilize the pretrained BERT model and GloVe word embeddings~\cite{pennington2014glove} to augment the input sentence via random word-level replacement.~MixKD~\cite{liang2020mixkd} adopts mixup~\cite{zhang2017mixup} and backtranslation~\cite{edunov2018understanding} in augmenting the text data to boost the performance of sentence classification models. Unlike these papers, we propose a novel search space for DA policies that supports stacking elementary augmentation operations such as EDA, mixup, and backtranslation. Thus, our considered DA module is similar to  AutoAugment~\cite{cubuk2019autoaugment}, except it is used for KD in NLP with different elementary operators. 
\section{Methodology}
\label{methodology}
Our study is structured around a configurable meta-distillation pipeline called \emph{Distiller}. Distiller  contains four configurable components, namely: a data augmentation policy $a(\cdot, \cdot)$, a layer mapping configuration of intermediate distillation $\{m_{i, j}\}$,  an intermediate distillation objective  $l^{\text{inter}}(\cdot,\cdot)$, and a prediction layer distillation objective $l^{\text{pred}}(\cdot,\cdot)$. Assume the teacher network $f^T$ has $M$ layers and the student network $f^S$ has $N$ layers. Then for a given  data/label pair $(x,y)$ sampled from the dataset $\mathcal{D}$, the student acquires knowledge from the teacher by minimizing the following objective: 
\begin{equation}
\label{eq:objective}
\begin{aligned}
    \mathcal{L} &=  \mathbb{E}_{\hat{x}, \hat{y} \sim a(x, y); x, y \sim \mathcal{D}} \sum_{i=1}^M \sum_{j=1}^N m_{i,j}l_{i,j}^{\text{inter}}(H_i^{\text{T}},H_j^{\text{S}})\\
    &+ \beta_1 l^{\text{pred}}(f^{T}(x),f^{S}(x)) + \gamma_1 l^{\text{pred}}(y, f^{S}(x))\\
    & + \beta_2 l^{\text{pred}}(f^{T}(\hat{x}),f^{S}(\hat{x}))+ \gamma_2 l^{\text{pred}}(\hat{y}, f^{S}(\hat{x})),
\end{aligned}
\end{equation}
Here $m_{i, j} \in [0, 1]$ represents the layer mapping weight between the $i$-th teacher layer and $j$-th student layer, $H_i^\text{T}, H_j^\text{S}$ are the $i$-th and the $j$-th hidden states of the teacher and the student (i.e.\ their intermediate representations at layers $i$ and $j$), $\beta_1$, $\beta_2$ control the strength of distilling from class probabilities produced by the teacher, and $\gamma_1$ and $\gamma_2$ control the strength of learning from ground truth data ($x,y$) and synthesized (augmented) data ($\hat{x},\hat{y}$). In Appendix, we illustrated how previous model distillation algorithms~\cite{jiao2019tinybert, li2020bert, liang2020mixkd} can be encompassed in the Distiller framework.
% We next discuss possible design choices for each component and also introduce the AutoDistiller.
\begin{figure}[tb!]
  \centering
  \includegraphics[width=0.4\textwidth]{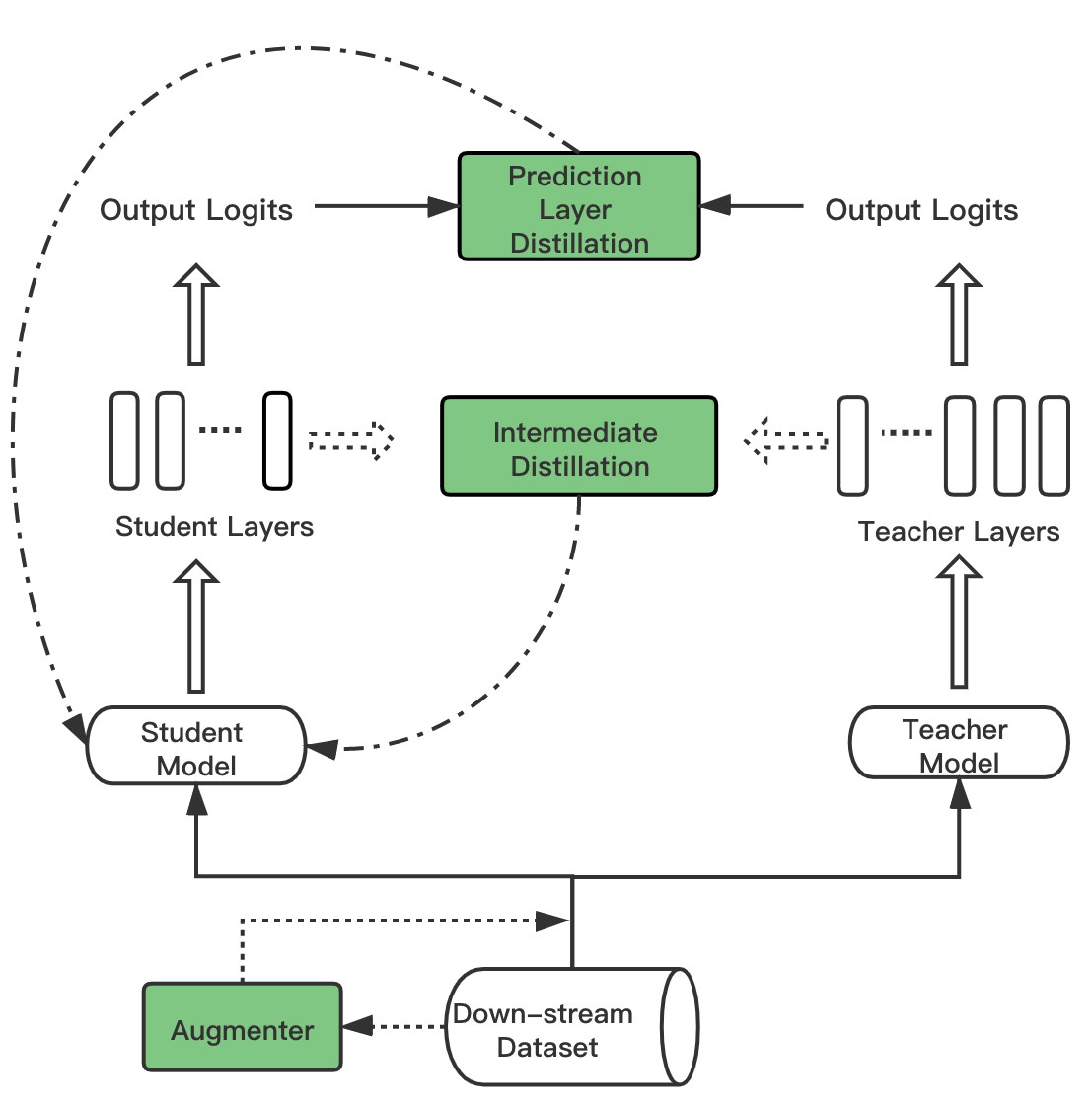}
  \caption{Overview of the \emph{Distiller} pipeline. All configurable components are colored.}
%   Augmenter is the data augmentation component. Intermediate distillation covers intermediate layer mapping strategy and intermediate objective. Prediction layer distillation calculates KD loss of the student output and teacher output.
\vspace{-1em}
\end{figure}
\subsection{Data Augmentation Policy}
\label{subsec:DA}
\SetKwInput{KwInitial}{Initialize}
\SetKwInput{KwParam}{Params}
\SetKwInput{KwIn}{Input}
\SetKwInput{KwOut}{Output}
\begin{algorithm}[tb!]
  \KwParam{A sequence of elementary data augmentation operations $\mathcal{G}$,
  $\forall \mathcal{G}_j \in \{\textit{CA}, \textit{RA}, \textit{BT},\textit{Mixup} \}$. }
%   \KwParam{$p_t$: the threshold probability; \\$n_k$: top k score tokens will be used for augmentation, only for $\textit{Contextual}$; $\lambda$: mixup ratio for $\textit{Mixup}$ augmentation}
  \KwIn{Training Dataset $\mathcal{D}_{\text{train}}$}
  \KwOut{Augmented dataset $\mathcal{D}_\text{synthesize}$}
  \BlankLine
  Initialize $\mathcal{D}_\text{synthesize} \leftarrow \{\}$ \\
    \ForEach {$\{x_i,y_i\} \in \mathcal{D}_{\text{train}} $}{
    \For{j $\leftarrow$ \text{1 to len}($\mathcal{G}$)}{
        $\hat{x}_i,\hat{y}_i=\mathcal{G}_j(x_i,y_i)$\;
        $x_i, y_i \leftarrow \hat{x}_i,\hat{y}_i $\;
     } 
    $\mathcal{D}_\text{synthesize}\leftarrow \mathcal{D}_\text{synthesize}\cup \{x_i,y_i\}$
    }
  \caption{Data Augmentation Policy}
  \label{algo:augmentation_alg}
\end{algorithm}
% In KD, training a student to achieve near performance to the teacher can be a tough task, especially when the teacher is much stronger than the student. 
A key challenge is the limited data available to train students in KD. 
This can be mitigated via Data Augmentation~(DA) to generate additional data samples. Unlike in supervised learning, where labels for synthetic augmented data may be unclear unless the augmentation is limited to truly benign perturbations, the labels for augmented data in KD are simply provided by the teacher which allows for more aggressive augmentation \cite{fakoor2020fast}. 
% opportunity for student to learn from a stronger teacher. 
Denoting the set of training samples of the down-stream task as $\mathcal{D}_{\text{train}}$, the augmenter $a(\cdot, \cdot)$ will stretch the distribution from $E_{x,y\sim {\mathcal{D}_{\text{train}}}}$ to $E_{\hat{x},\hat{y}\sim {a(x,y)},x,y\sim {\mathcal{D}_{\text{train}}}}$. We consider various elementary DA operations including: 1) MLM-based contextual augmentation (CA) , 2) random augmentation (RA), 3) backtranslation (BT) and 4) mixup. 
The search space of possible augmentations in \emph{Distiller} is constructed by stacking these four elementary operations in an arbitrary order, as detailed in Algorithm~\ref{algo:augmentation_alg}. 

For contextual augmentation, we use the pretrained BERT model to do word level replacement by filling in randomly masked tokens. As in EDA  \cite{wei2019eda}, our random augmentation randomly swaps words in the sentence or replaces words with their synonyms. For backtranslation, we translate the sentence from one language (in this paper, English) to another language (in this paper, German) and then translate it back. 
% This can produce augmented sentences that have similar semantic meaning but very different structure. 
% In our implementation, we translate from English to German and then translate German to English.
%In our implementation since we focus on English datasets, we implement English to German and translate 
Additionally, mixup can be used to synthesize augmented training samples.  First proposed for image classification~\cite{zhang2017mixup}, mixup  constructs a synthetic training example via the weighted average of two samples (including the labels) drawn at random from the training data. To use it in NLP, \citet{guo2019augmenting,liang2020mixkd} applied mixup on the word embeddings at each sentence position $x_{i,t}$ with $\lambda \in [0,1]$ as the mixing-ratio for a particular pair of examples $x_i, x_j$:
\begin{equation*}
    \hat{x}_{i,t}=\lambda x_{i,t} + (1-\lambda) x_{j,t},~ \hat{y}_i=\lambda y_i + (1-\lambda) y_j,
\end{equation*}
Here $\lambda$ is typically randomly drawn from a \textit{Uniform} or \textit{Beta} distribution for each pair, $y_i,y_j$ are labels in one-hot vector format, and $(\hat{x}_{i},\hat{y}_i)$ denotes the new augmented sample. To apply mixup for sentence tagging tasks, in which each token has its own label, we propose calculating the weighted combination of the ground-truth target at each location $t$ as the new target: 
\begin{equation*}
    \hat{x}_{i,t}=\lambda x_{i,t} + (1-\lambda) x_{j,t}, ~ \hat{y}_{i,t}=\lambda y_{i,t} + (1-\lambda) y_{j,t},\
\end{equation*}
% Here $y_{i,t}$ is the probability distribution for position $t$ to be selected as start or end position of the answer span. 

\subsection{Prediction Layer Distillation}
In traditional KD, the student network learns from the output logits of the teacher network, adopting these as soft labels for the student's training data \cite{hinton2015distilling}.  
Here we penalize the discrepancy between the outputs of student vs.\ teacher via:
\begin{equation}
    \mathcal{L}_{\text{pred}}=l^{\text{pred}}(f^T(x), f^S(x)),
\end{equation}
where $l^{\text{pred}}(\cdot,\cdot)$ is the KD loss component whose search space in this work includes either: softmax Cross-Entropy (CE) or Mean Squared Error (MSE).
\vspace{-2em}
\subsection{Intermediate Representation Distillation}
To ensure knowledge is sufficiently transferred, we can allow the student to learn from intermediate layers of the teacher rather than only the latter's output predictions by minimizing discrepancies between selected layers from the teacher and the student.
% \JM{add intermediate distillation citations here}.
These high-dimensional intermediate layer representations constitute a much richer information-dense signal than is available in the low-dimensional predictions from the output layer. 
% BERT-PKD \cite{sun2019patient} shows that this intermediate distillation scheme enables the student to \emph{patiently} learns the rich information in the teacher's hidden layers. 
As teacher and student usually have different number of layers and hidden-state dimensionalities, it is not clear how to map teacher layers to student layers ($m_{i,j}$) and how to measure the discrepancy between their hidden states ($l_{i,j}^{\text{inter}}$). Previous works proposed various discrepancy measures (or loss functions) for intermediate distillation, including:  Cross-Entropy (CE), Mean Squared Error (MSE), L2 distance, Cosine Similarity (Cos), and  Patient Knowledge Distillation (PKD)~\cite{sun2019patient}.
For these objectives, we establish the following result (the proof is relegated to the Appendix). 
% Based on the recent advancement of Mutual Information (MI) estimation, we proved that optimizing the aforementioned loss functions can be viewed as maximizing different lower bounds of MI, as described in the following theorem.

\begin{theorem}

Minimizing MSE, L2, or PKD loss, and maximizing cosine similarity between two random variables $X$,$Y$ are equivalent to maximizing lower bounds of the mutual information $I(X;Y)$. 
\label{theorem:unify}
\end{theorem} 
% The proof of Theorem~\ref{theorem:unify} is relegated to the Appendix. 
In our KD setting, $X$ and $Y$ correspond to the hidden state representations of our student and teacher model (for random training examples), respectively. 
Inspired by this result, we can use any lower bounds of MI as an intermediate objective function in KD. 
In particular, we consider the multisample MI lower bound of  \citet{poole2019variational}, which estimates $I(X;Y)$ given the sample $x, y$ from $p(x, y)$ and another $K$ additional IID samples $z_{1:{K}}$ that are drawn from a distribution independent from $X$ and $Y$:

\begin{scriptsize}
\centering
\begin{align*}
     I(X;Y) &\geq E_{p(x, z_{1:{K}})p(y|x)} \left[\log \frac{e^{f(x,y)}}{\alpha m (y;x, z_{1:{K}})+(1-\alpha)q(y)} \right] \\
    & - E_{p(x, z_{1:{K}})p(y)} \left[\log \frac{e^{f(x,y)}}{\alpha m (y;x, z_{1:{K}})+(1-\alpha)q(y)} \right] + 1\\
    & \triangleq I_{\alpha}. \numberthis
\end{align*}
\end{scriptsize}
% \JM{$X_{2:K}$ seems to be missing from this formula and should be discussed more. Also would be good to use $Z$ instead of $X_2$, so that way you don't need to introduce the weird notation $X_1$ and can just use $X$ instead.}
\vspace{-1em}

In $I_\alpha$, $f(\cdot,\cdot)$ and $q(\cdot)$ are critic functions for approximating unknown densities  and $m(\cdot,\cdot)$ is a Monte-Carlo estimate of the partition function that appears in MI calculations. Typically, the space $z$ and the sample $x,y$ are from the same minibatch while training, that is $K+1$ equals to the minibatch size. $I_\alpha$ can flexibly trade off bias and variance, since increasing $\alpha \in [0, 1]$ will reduce the variance of the estimator while increasing its bias. We propose to use $I_\alpha$ as an objective for intermediate distillation and call it MI-$\alpha$. Our implementation leverages a Transformer encoder~\cite{vaswani2017attention} to learn $f(\cdot,\cdot)$ and $q(\cdot)$. To our knowledge, this is the first attempt to utilize complex NN architectures for critic functions in MI estimation; typically only shallow multilayer perceptrons (MLPs) are used~\cite{tschannen2019mutual}. Our  experiments (Table \ref{tab:MI-critics} in Appendix) reveal that Transformer produces a better critic function than MLP.

Note that for intermediate distillation, objectives like MSE attempt to ensure the teacher and student representations take matching values, whereas objectives like MI (and tighter bounds thereof) merely attempt to ensure the information in the teacher representation is also captured in the student representation. The latter aim is conceptually better suited for KD, particularly in settings where the student's architecture differs from the teacher (e.g.\ it is more compact), in which case forcing intermediate student representations to take the exact same values as teacher representations seems overly stringent and unnecessary for a good student (it may even be harmful for  tiny student networks that lack the capacity to learn the same function composition used by the teacher). We emphasize that a high MI between student and teacher representations suffices for the teacher's prediction to be approximately recovered from the student's intermediate representation (assuming the teacher uses deterministic output layers as is standard in today's NLP models). Given that high MI suffices for the student to match the teacher, we expect tighter MI bounds like MI-$\alpha$ can outperform looser bounds like MSE that impose additional requirements on the student's intermediate representations beyond just their information content.

\subsubsection{Layer Mapping Strategy}
We investigate three intermediate layer mapping strategies: 1) Skip: the student learns from every $\floor{M/N}$ layer of the
% \todo{denote layer mapping strategy with $m_{i,j}$ or a function $n=g(m)$}
teacher, i.e., $m_{i,j}=1$ when $j=i\times \floor{M/N}$; 2) Last: the student learns from the last $k$ layers of the teacher, i.e., $m_{i,j}=1$ when $j=i+M - N$; and 3) EMD: a many-to-many learned layer mapping strategy~\cite{li2020bert} based on Earth Mover's Distance. In the \emph{Distiller} pipeline, the intermediate loss with EMD mapping can be denoted as:
\vspace{-1em}
\begin{equation}
    \mathcal{L}_{\text{EMD}}(H^S_{1:N},H^T_{1:M})=\frac{\sum_{i=1}^{M}\sum_{j=1}^{N}w_{i,j}^H d_{i,j}^H}{\sum_{i=1}^{M}\sum_{j=1}^{N}w_{i,j}^H},
\end{equation}
where $D^H=[d_{i,j}^H]$ is a distance matrix representing the cost of transferring the hidden states knowledge from $H^T$ to $H^S$. And $W^H=[w_{i,j}^H]$ is the mapping flow matrix which is learned by minimizing the cumulative cost required to transfer knowledge from $H^T$ to $H^S$. In \emph{Distiller}, the distance matrix is calculated via intermediate objective function: $d_{i,j}^H=l^{\text{inter}}(H_i^S,H_j^T)$.  

% {\jonas{TODO:Here is where we could say something like this: Distiller is a generic meta framework that encompasses various KD pipelines used in previous work. For example, Distiller with the following configurations corresponds to the KD pipeline used in each of the cited works: 
% $l^{\text{pred}} =$ CE, $l^{\text{inter}}=$ MSE, $m_{i,j} = $ Skip, $a = $ CA  \cite{jiao2019tinybert}; 
% $l^{\text{pred}} =$ CE, $l^{\text{inter}}=$ MSE, $m_{i,j} = $ EMD \cite{li2020bert}; 
% $l^{\text{pred}} =$ CE, $a = $ Mixup  \cite{liang2020mixkd}.
% }}

\subsection{AutoDistiller}

Our experiments indicate that the best distillation algorithm varies among datasets/tasks (see in particular the top-5 configurations listed in Table~\ref{tab:top5} in Appendix). This inspires us to train a prediction model that recommends a good KD pipeline given a dataset. To represent the distillation performance across datasets which are evaluated on different metrics, we define distillation ratio as the fraction of the teacher's performance achieved by the student and use it as a general score for distillation performance. Then the prediction model can be trained to predict the distillation ratio based on features of the dataset/task as well as the features of each candidate distillation pipeline. Here we train our \emph{AutoDistiller} performance prediction model via AutoGluon-Tabular, a simple AutoML tool for supervised learning~\cite{erickson2020autogluon}. To the best of our knowledge, our proposed method is the first attempt towards automated KD in NLP.

\section{Experimental Setup}
\label{exps}
To study the importance of each component described in the previous section, we randomly sample \emph{Distiller} configurations in the designed search space while fixing the optimizer and other unrelated hyper-parameters. We apply each sampled distillation configuration on a diverse set of NLP tasks and different teacher/student architectures.
All experiments are evaluated on GLUE~\cite{wang2018glue} and SQuAD v1.1~\cite{rajpurkar2016squad} that contain classification, regression, and sentence tagging tasks. Here, we view the question answering problem in SQuAD v1.1 as finding the correct answer span from the given context, which is essentially a sentence tagging task. We adopt the same metrics for these tasks as in the original papers~\cite{wang2018glue, rajpurkar2016squad}.

Since \citet{turc2019well} finds initializing students with pretrained weights is better for distillation, we initialize student models with either weights obtained from task-agnostic  distillation~\cite{jiao2019tinybert} or pretrained from scratch~\cite{turc2019well}.~Three different pretrained models $\text{BERT}_\text{BASE}$~\cite{devlin2018bert}, $\text{RoBERTa}_\text{LARGE}$~\cite{liu2019roberta} and $\text{ELECTRA}_\text{LARGE}$~\cite{clark2020electra} are considered as teacher models in our experiments after task-specific fine-tuning.~As student models, we consider options like $\text{TinyBERT}_4$, $\text{ELECTRA}_\text{SMALL}$, as well as other models detailed in Table~\ref{tab:teacher_student_architecture} in Appendix.
% For evaluating the performance of MI-$\alpha$, we initialize our student model with the 4-layer general distillation model provided by TinyBERT\cite{jiao2019tinybert}, and we finetune BERT model on specific datasets as teacher model. 

Existing implementations of data augmentation in KD for NLP  generally first generate an augmented dataset that is $K$ times larger than the original one and then apply the distillation algorithm over the augmented dataset. Such implementation separates the process of DA from KD, leading to a  time/storage-consuming and inflexible KD pipeline. In \emph{Distiller}, we instead apply DA dynamically during training. In addition, we use the teacher network to compute the soft label $\hat{y}$ assigned to any augmented sample $\hat{x}$.

% This sentence is redundant as it is already described in MI-alpha section: We implement MI-$\alpha$ using  various neural-network architectures as critic functions (details in Appendix \ref{sec:app:com-config}).

To analyze the importance of different components in \emph{Distiller}, we adopted fANOVA~\cite{hutter2014efficient}, an algorithm for quantifying the importance of individual hyper-parameters as well as their interactions in determining down-stream performance. We use fANOVA to evaluate the importance of the four components in \emph{Distiller} as well as their pairwise combinations: data augmentation, intermediate distillation objective, layer mapping strategy, and prediction layer distillation objective.

\begin{table*}[tb!]
\centering
\caption{Comparison of evaluation results on GLUE test set.  $\text{BERT}_{\text{BASE}}$~\text{(G)} and $\text{BERT}_{\text{BASE}}$~\text{(T)} indicate the fine-tuned $\text{BERT}_{\text{BASE}}$ from \cite{devlin2018bert} and the teacher model trained by ourselves, respectively. $\text{BERT-EMD}_4$ and MI-$\alpha$ are both initialized from $\text{TinyBERT}_4$, the difference is that $\text{BERT-EMD}_4$ is trained with ``EMD'' as intermediate layer mapping strategy and MSE as intermediate loss, our MI-$\alpha$ model is trained with ``Skip'' as intermediate layer mapping strategy and MI-$\alpha$ as intermediate loss.}
\resizebox{0.95\linewidth}{!}{%
\begin{tabular}{ll|cccccccccc}
\toprule
Model       & \#Params & MNLI-m & MNLI-mm & QQP           & QNLI   & SST-2 & CoLA   & MRPC   & RTE    & STS-B  & AVG   \\
            &        & (393k) & (393k)  & (364k)        & (108k) & (67k) & (8.5k) & (3.5k) & (2.5k) & (5.7k) &       \\ \hline
$\text{BERT}_{\text{BASE}}$~\text{(G)} & 110M   & 84.6   & 83.4    & 71.2          & 90.5   & 93.5  & 52.1   & 88.9   & 66.4   & 85.8   & 79.6  \\
$\text{BERT}_{\text{BASE}}$~\text{(T)} & 110M   & 84.5   & 83.6    & 71.7          & 90.9   & 93.4  & 49.3   & 87.0   & 67.3   & 84.7   & 79.2 \\ \hline
$\text{BERT-PKD}_4$~\cite{sun2019patient}  & 52M    & 79.9   & 79.3    & \textbf{70.2} & 85.1   & 89.4  & 24.8   & 82.6   & 62.3   & 82.3   & 72.9 \\
$\text{BERT-EMD}_4$~\cite{li2020bert} & 14M & \textbf{82.1} & \textbf{80.6} & 69.3 & 87.2          & 91.0            & 25.6 & \textbf{87.6} & 66.2          & 82.3 & 74.7          \\
MI-$\alpha$ ($\alpha=0.9$, ours)  & 14M & 81.9          & \textbf{80.6} & 69.8 & \textbf{87.4} & \textbf{91.5} & \textbf{25.9}          & 87.0          & \textbf{67.4} & \textbf{84.0} & \textbf{75.1} \\
\bottomrule
\end{tabular}}
\label{tab:mi-alpha}
\end{table*}
\section{Experimental Results}
\label{sec:bibtex}
Under the previously described experimental setup, we conducted a random search over \emph{Distiller} configurations on each dataset and collected more than 1300 data points in total. Each collected data point contains a particular \emph{Distiller} configuration, the dataset/task, the teacher/student architectures, and the final performance of the student model. Analyzing the data reveals three major findings:
\vspace{-0.5em}
\begin{enumerate}
    \item Design of the intermediate distillation module is the most important choice among all factors studied in Distiller.
    \vspace{-0.5em}
    \item Among different loss functions for intermediate distillation, MI-$\alpha$ performs the best.
    \vspace{-0.5em}
    \item DA provides a large boost when the dataset or the student model is small.
    \vspace{-0.5em}
\end{enumerate}
Additionally, we observe that the best distillation policy varies among datasets/tasks as shown in Table~\ref{tab:top5} in Appendix. Thus we train a meta-learning model, \emph{AutoDistiller}, that can recommend a good distillation policy on any new NLP dataset based on which configurations tended to work well for similar datasets in our study.

\subsection{Importance of Distiller Components}
\begin{figure}[!tb]
    \vspace{-2em}
    \centering
    \includegraphics[width=0.8\columnwidth]{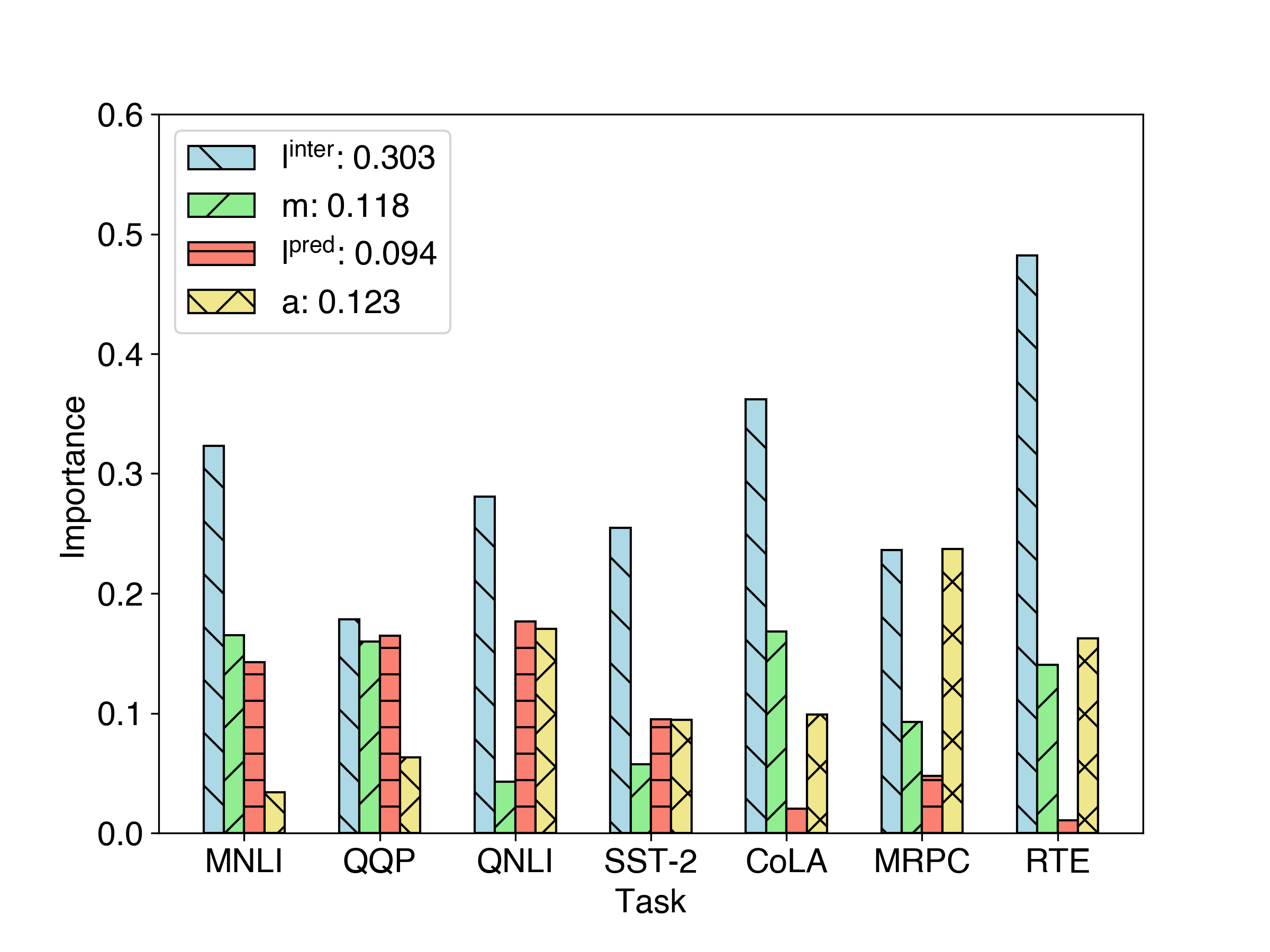}
    % \caption{Importance of single hyper-parameters.}
%     \begin{subfigure}[b]{0.8\columnwidth}
% %    \begin{minipage}{6cm}
%     \centering
%     \vspace{-1em}
%     \includegraphics[width=1.0\columnwidth]{pics/fanova_double_wo_stsb.png}
% %    \end{minipage}
%     \caption{Importance of interactions between hyper-parameters.}
%     \end{subfigure}
    \caption{As assessed via fANOVA, we report the individual importance of the four Distiller components in this figure and importance of interactions between any two of the four components in Figure \ref{fig:fanova-interactions} in Appendix. Four components are: $l^{\text{inter}}$ for intermediate distillation objective, $l^{\text{pred}}$ for prediction layer objective, $a$ for data augmentation and $m$ for layer mapping strategy. Average importance for each component (across tasks) is listed in the legend.}
    \label{fig:fanova}
    \vspace{-1em}
\end{figure}

 Figure~\ref{fig:fanova} illustrates that the objective function for intermediate distillation $l^{\text{pred}}$ has the highest individual importance out of all components in \emph{Distiller}, and the combination of the intermediate distillation objective and layer mapping strategy has the highest joint importance. Thus these are the components one should most critically focus on when selecting a particular KD pipeline. One hypothetical explanation is that the teacher can provide token-level supervision to the student via intermediate distillation, which can better guide the learning process of the student. Our finding is also consistent with the previous observations~\cite{sun2019patient,li2020bert}.

\subsection{MI-$\alpha$ for Intermediate Distillation}
\begin{figure}[tb!]
% \todo{Split different colors in case mono print}
  \centering
  \includegraphics[width=0.8\columnwidth]{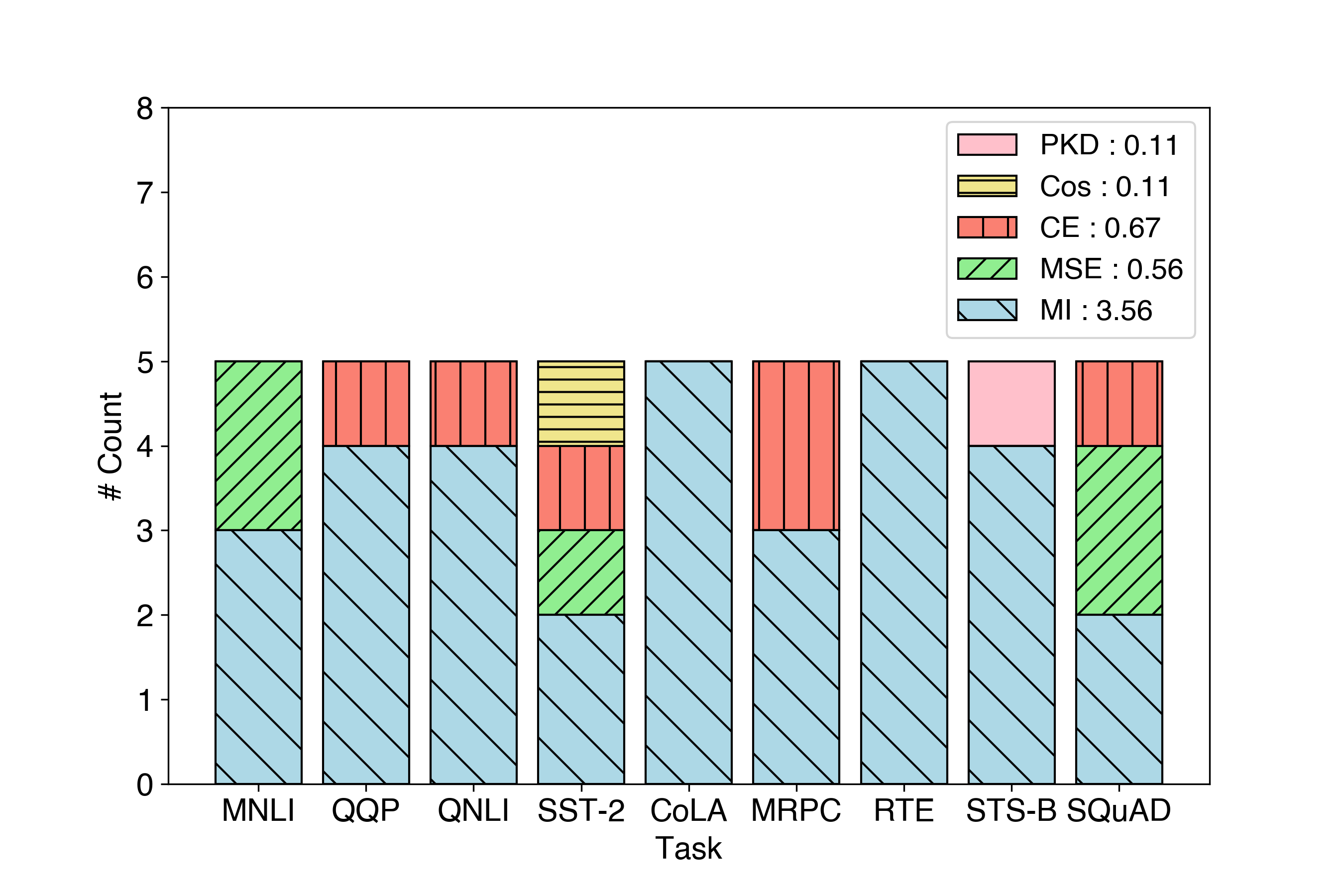}
  \caption{Intermediate objective functions used in the top-5 performing KD configurations on each dataset. The average count of each objective function is listed in the legend. Configurations are detailed in Appendix.}
  \label{fig:top5}
  \vspace{-1em}
\end{figure}

We submitted our MI-$\alpha$ model predictions to the official GLUE leaderboard to obtain test set results and also report the average scores over all tasks (the ``AVG'' column) as summarized in Table \ref{tab:mi-alpha}. The results show that the student model distilled via the MI-$\alpha$ objective function outperforms previous student models distilled via MSE or PKD loss.
% \todo{May compare to other baselines which can control variables to MI-$\alpha$}
To further verify the effectiveness of MI-$\alpha$, we compare how different choices of $l^{\text{inter}}$ affect the distillation performance. In detail, we first pick the top 5 best \emph{Distiller} strategies according to the evaluation scores for every task and then count how many times each intermediate objective function appears in these strategies. Figure \ref{fig:top5} shows that MI-$\alpha$ appears more frequently than all other objectives on both classification and regression tasks. And the results for SQuAD v1.1 in Table~\ref{tab:mi-alpha-squad} indicate the MI-$\alpha$ also works well for sentence tagging. \\

\begin{table}[tb!]
\centering
\caption{Ablation study of distillation performance on SQuAD v1.1 dev set. The first line shows the performance of the $\text{BERT}_\text{BASE}$ teacher. $\text{ELECTRA}_\text{SMALL}$, $\text{TinyBERT}_6$ and $\text{TinyBERT}_4$ are three student networks. $\text{ELECTRA}_{\text{SMALL}}$~(FT) means to fine-tune without KD. ${\text{TinyBERT}_6}^\dagger$ and ${\text{TinyBERT}_4}^\dagger$ are results obtained from~\citet{jiao2019tinybert}. Models that end with ``(MSE)'' are trained with the MSE loss. ``+ MI-$\alpha$'' means to distill the student with MI-$\alpha$ ($\alpha$=0.9) as the intermediate loss function. ``+ mixup'' means to further apply the mixup augmentation.}
\resizebox{0.75\linewidth}{!}{%
\begin{tabular}{l|cll}
\toprule
Model & \multicolumn{2}{l}{\textbf{SQuAD v1.1}} \\
 & EM & F1 \\
\hline
$\text{BERT}_{\text{BASE}}$~\text{(T)} &  80.9 & 88.2 \\
\hline
%Student models with MI-$\alpha$ as intermediate loss \\ \hline
$\text{ELECTRA}_{\text{SMALL}}$~(FT) &  75.3 & 83.5 \\
$\text{ELECTRA}_{\text{SMALL}}$ (MSE)  & 79.2 & 86.8 \\
+ MI-$\alpha$  & 79.0 & 86.8 \\
+ mixup, MSE & 80.1 & 87.4 \\ 
+ mixup, MI-$\alpha$ & \textbf{80.2} & \textbf{87.6} \\ 
\hline
${\text{TinyBERT}_6}^\dagger$~\cite{jiao2019tinybert} &  79.7 & 87.5 \\
$\text{TinyBERT}_6$ (MSE) & 77.8 & 85.5 \\
+ MI-$\alpha$ & 80.0 & 87.8 \\
+ mixup, MSE & 78.6 & 86.2 \\ 
+ mixup, MI-$\alpha$ & \textbf{81.1} & \textbf{88.6} \\ 
\hline
${\text{TinyBERT}_4}^\dagger$~\cite{jiao2019tinybert} &  \textbf{72.7} & 82.1 \\
$\text{TinyBERT}_4$ (MSE) & 72.7 & 81.7  \\
+ MI-$\alpha$ & 71.7 & 81.2 \\
+ mixup, MSE & 72.4 & 81.4\\ 
+ mixup, MI-$\alpha$ & 71.9 & \textbf{82.5}\\ 
\bottomrule
\end{tabular}}
\label{tab:mi-alpha-squad}
\end{table}
\vspace{-2em}
% We suspect that this is because the representation power of the 2-layer MLP critic functions in MI-$\alpha$ is too little to catch the feature for every span in a sentence. 
% \JM{this meaning is unclear: too little to catch the feature for every span in a sentence}
\subsection{Benefits of Data Augmentation}
Data augmentation in KD provides the student additional opportunities to learn from the teacher, especially for datasets of limited size. Thus, our experiments investigate the effect of DA on four data-limited tasks: CoLA, MRPC, RTE and STS-B. 
We also study whether students with different architectures/sizes benefit dissimilarly from DA. Table \ref{tab:aug} demonstrates that DA generally provides a boost to student performance and is especially beneficial for small models ($\text{BERT}_\text{MINI}$ and $\text{BERT}_\text{TINY}$).

% For ``smarter'' students such as $\text{ELECTRA}_\text{SMALL}$ (pretrained on a better-designed objective) and $\text{TinyBERT}_4$ (task-agnostic distilled from $\text{BERT}_\text{BASE}$), they get comparably less boost from DA.
%One possibility is that these models have already seen converge more data than those undertrained models in the phase of pre-training (or general distillation), so they benefit less through DA.

\begin{table}[tb!]
\centering
\caption{Student performance with(out) augmentation (augmenter initialized as CA+RA+mixup). We report the relative improvement for rows starting with ``+ aug''.}
\resizebox{1.0\linewidth}{!}{%
\begin{tabular}{lc|ccccccc}
\toprule
Model & \#Params & CoLA & MRPC & RTE & STS-B & AVG \\
 &  & mcc & f1/acc & acc & spearman/pearson & \multicolumn{1}{l}{} \\ \hline
$\text{BERT}_{\text{BASE}}~\text{(T)}$ & 110M & 55.0 & 89.6/85.0 & 65.0 & 88.4/88.6 & 78.6 \\ \hline
$\text{TinyBERT}_{\text{6}}$ & 67M & 51.3 & \textbf{92.5}/\textbf{89.7} & \textbf{75.5} & 89.6/89.8 & \textbf{81.4} \\
+ aug &  & \textbf{+0.1} & -1.1/-1.8 & -3.3 & \textbf{+0.2}/\textbf{+0.2} & -1.0 \\ \hline
$\text{BERT}_{\text{MEDIUM}}$ & 41M & 44.1 & \textbf{89.3}/\textbf{84.8} & 65.3 & 88.3/88.6 & 76.7 \\
+ aug &  & \textbf{+5.3} & -0.4/-0.7 & \textbf{+4.4} & \textbf{+0.6}/\textbf{+0.5} & \textbf{+1.6} \\ \hline
$\text{BERT}_{\text{SMALL}}$ & 29M & 37.4 & 86.8/80.6 & 64.6 & 87.7/88.0 & 74.2 \\
+ aug &  & \textbf{+5.0} & \textbf{+0.1}/\textbf{+0.8} & \textbf{+0.4} & \textbf{+0.3}/\textbf{+0.2} & \textbf{+1.1} \\ \hline
$\text{TinyBERT}_{\text{4}}$ & 14M & 23.6 & 88.9/\textbf{83.8} & 67.1 & 88.0/88.1 & 73.3 \\
+ aug &  & \textbf{+7.9} & \textbf{+0.3}/\textbf{+0.0} & \textbf{+2.2} & \textbf{+0.7}/\textbf{+0.7} & \textbf{+2.0} \\ \hline
$\text{ELECTRA}_{\text{SMALL}}$ & 14M & 42.8 & 88.3/83.8 & 66.4 & 87.4/87.5 & 76.0 \\
+ aug &  & \textbf{+16.2} & \textbf{+3.4}/\textbf{+3.7} & \textbf{+1.8} & \textbf{+1.0}/\textbf{+1.0} & \textbf{+4.5} \\ \hline
$\text{BERT}_{\text{MINI}}$ & 11M & 11.2 & \textbf{86.1}/\textbf{80.1} & 62.8 & 87.1/\textbf{87.2} & 69.1 \\
+ aug &  & \textbf{+23.2} & \textbf{+0.0}/-0.1 & \textbf{+3.3} & \textbf{+0.2}/\textbf{+0.0} & \textbf{+4.4} \\ \hline
$\text{BERT}_{\text{TINY}}$ & 4M & 6.0 & 83.2/73.3 & 60.0 & 84.0/83.6 & 65.0 \\
+ aug &  & \textbf{+6.6} & \textbf{+1.7}/\textbf{+3.7} & \textbf{+4.3} & \textbf{+0.1}/\textbf{+0.7} & \textbf{+2.9} \\ \bottomrule 
\end{tabular}}
% In this experiment, we initialize our augmenter as: CA + RA + mixup. Results are evaluated on GLUE dev set and best results are in bold.}
\label{tab:aug}
\end{table}

\vspace{-0.5em}
\subsection{Performance of AutoDistiller}
Recall that we train our \emph{AutoDistiller} performance prediction model on the previously collected experimental results via AutoGluon-Tabular. Once trained,  \emph{AutoDistiller} can recommend distillation pipelines for any down-stream dataset/task by fixing the dataset/task features and searching for configurations that maximize the predicted distillation ratio. \emph{AutoDistiller} operates on features that represent the dataset domain, the task type, and the task complexity, which are detailed in Appendix. We evaluate the performance of \emph{AutoDistiller} on the 8 GLUE datasets via a leave-one-dataset-out cross-validation protocol. Figure~\ref{fig:leaveoneout} in Appendix shows that \emph{AutoDistiller} achieves positive Spearman's correlation coefficients for most datasets. \\
% \JM{ranking coefficients is not a well-known technical term, so should be defined or a more precise term used instead.}
% \begin{figure}[!tb]
%   \centering
%   \begin{subfigure}[b]{0.5\columnwidth}
% 	\centering          
% 	\includegraphics[scale=0.5]{pics/leave_one_out_predictions_100d.jpg}
%   \end{subfigure}
%   \hfill
%   \begin{subfigure}[b]{0.45\columnwidth}
% 	\centering         
% 	\includegraphics[scale=0.3]{pics/leave_one_out_predictions2_100d.jpg}
%   \end{subfigure}
%   \caption{AutoDistiller evaluations on GLUE. We use a leave-one-out estimate, holding out a dataset from GLUE for the cross validation purpose. Spearman's rank correlation of AutoDistiller predicted distillation ratio against true distillation ratio is listed in legend.
%   }
\vspace{-1em}

\begin{figure}[!tb]
    \vspace{-1em}
    \centering
    \includegraphics[width=0.35\textwidth]{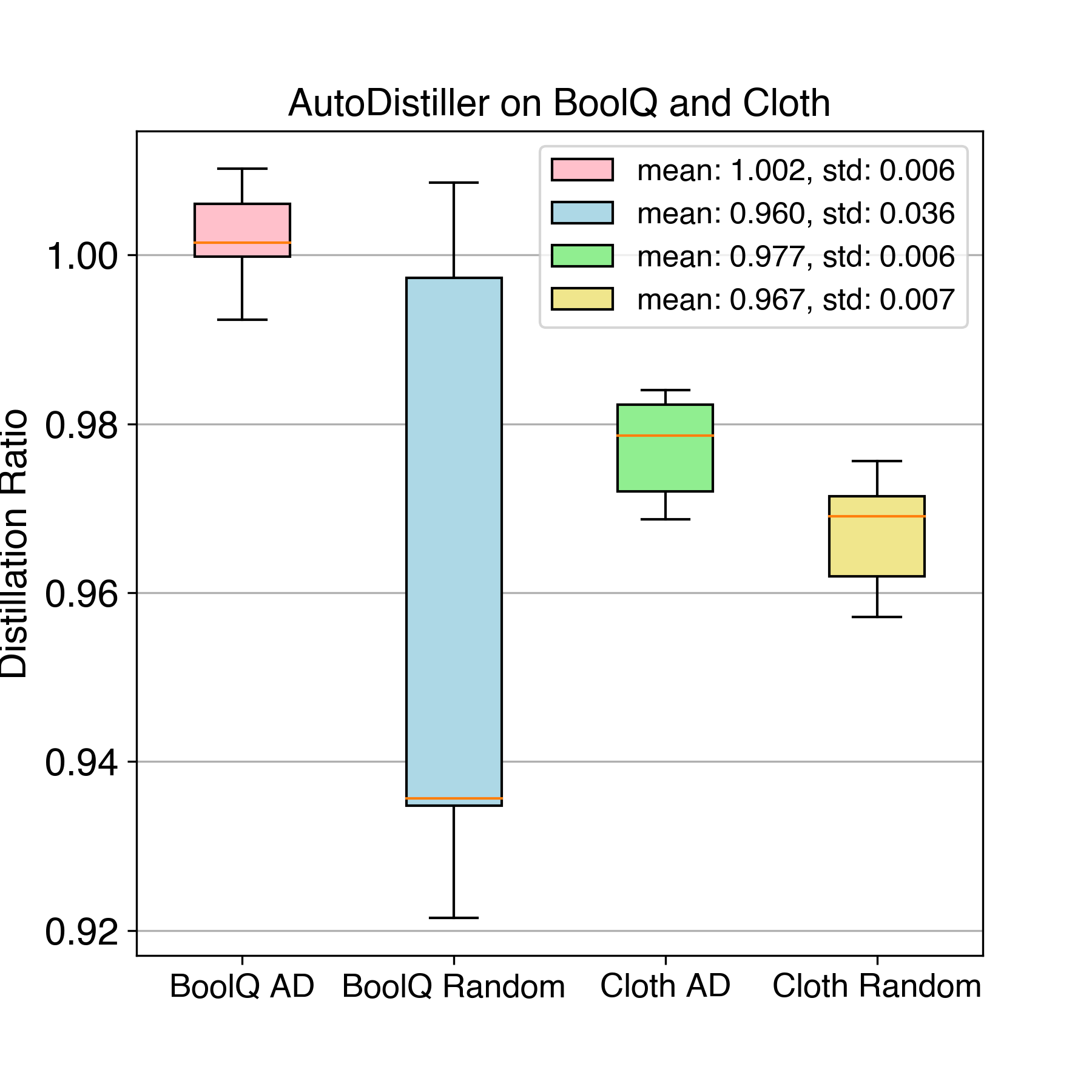}
    \caption{Distillation ratio of \emph{AutoDistiller} (AD) top-5 KD strategies vs.\ 5 randomly selected strategies, for a fine-tuned $\text{BERT}_\text{BASE}$ teacher and $\text{TinyBERT}_4$ student. Higher ratio indicates better distillation performance. Mean and standard deviation of the four groups of ratios are listed in the legend.}
    \label{fig:autodistiller}
    % \vspace{-1em}
    \vspace{-1em}
\end{figure}

Finally, we applied \emph{AutoDistiller} on two unseen datasets ``BoolQ''~\cite{wang2019superglue} and ``cloth''~\cite{shi2021multimodal} not considered in our previous experiments. We compared the distillation ratio obtained by each of the top-$N$ strategies suggested by \emph{AutoDistiller} with the distillation ratio from each of $N$ randomly selected strategies. The best strategy suggested by \emph{AutoDistiller} achieves accuracy of $74.2$ on ``BoolQ'' and $70.1$ on ``cloth'', close or superior to the teacher performance ($73.4$ on ``BoolQ'' and ``71.2'' on cloth). Figure \ref{fig:autodistiller} shows that \emph{AutoDistiller} significantly outperforms random search, indicating its promise for automated KD.
% The random selection rule is: randomly sample N configurations from Distiller search space and keep the configuration with best prediction. 
\section{Conclusion}
We provided a systematic study of KD algorithms in NLP to understand the importance of different components in the KD pipeline for various NLP tasks. 
Using data collected from our study, we fit a \emph{AutoDistiller} to predict student performance under each KD pipeline based on dataset features, which 
% proved to outperform random sampled distillation pipelines.
is helpful for automatically selecting which KD pipeline to use for a new  dataset. 
% Our study reveals that the intermediate distillation module's design has an outsized effect on student performance.
Here we unified the existing intermediate distillation objectives as maximizing the lower bounds of MI, leading to a new MI-$\alpha$ objective based on tighter bounds, which performs better on many datasets. 
% In future work, we will generalize our study to KD on multimodal datasets and also  extend to apply MI-$\alpha$ in other KD tasks.
\clearpage
% \balance{}
\bibliographystyle{acl_natbib}
\bibliography{distiller}

\begin{thebibliography}{41}
\expandafter\ifx\csname natexlab\endcsname\relax\def\natexlab#1{#1}\fi

\bibitem[{Blei et~al.(2017)Blei, Kucukelbir, and
  McAuliffe}]{blei2017variational}
David~M Blei, Alp Kucukelbir, and Jon~D McAuliffe. 2017.
\newblock Variational inference: A review for statisticians.
\newblock \emph{Journal of the American Statistical Association},
  112(518):859--877.

\bibitem[{Brown et~al.(2020)Brown, Mann, Ryder, Subbiah, Kaplan, Dhariwal,
  Neelakantan, Shyam, Sastry, Askell, Agarwal, Herbert-Voss, Krueger, Henighan,
  Child, Ramesh, Ziegler, Wu, Winter, Hesse, Chen, Sigler, Litwin, Gray, Chess,
  Clark, Berner, McCandlish, Radford, Sutskever, and Amodei}]{gpt3}
Tom Brown, Benjamin Mann, Nick Ryder, Melanie Subbiah, Jared~D Kaplan, Prafulla
  Dhariwal, Arvind Neelakantan, Pranav Shyam, Girish Sastry, Amanda Askell,
  Sandhini Agarwal, Ariel Herbert-Voss, Gretchen Krueger, Tom Henighan, Rewon
  Child, Aditya Ramesh, Daniel Ziegler, Jeffrey Wu, Clemens Winter, Chris
  Hesse, Mark Chen, Eric Sigler, Mateusz Litwin, Scott Gray, Benjamin Chess,
  Jack Clark, Christopher Berner, Sam McCandlish, Alec Radford, Ilya Sutskever,
  and Dario Amodei. 2020.
\newblock Language models are few-shot learners.
\newblock In \emph{Advances in Neural Information Processing Systems}.

\bibitem[{Buciluǎ et~al.(2006)Buciluǎ, Caruana, and
  Niculescu-Mizil}]{bucilua2006model}
Cristian Buciluǎ, Rich Caruana, and Alexandru Niculescu-Mizil. 2006.
\newblock Model compression.
\newblock In \emph{Proceedings of the 12th ACM SIGKDD international conference
  on Knowledge discovery and data mining}, pages 535--541.

\bibitem[{Chen et~al.(2020)Chen, Li, Qiu, Wang, Li, Ding, Deng, Huang, Lin, and
  Zhou}]{chen2020adabert}
Daoyuan Chen, Yaliang Li, Minghui Qiu, Zhen Wang, Bofang Li, Bolin Ding, Hongbo
  Deng, Jun Huang, Wei Lin, and Jingren Zhou. 2020.
\newblock Adabert: Task-adaptive bert compression with differentiable neural
  architecture search.
\newblock In \emph{IJCAI}.

\bibitem[{Clark et~al.(2020)Clark, Luong, Le, and Manning}]{clark2020electra}
Kevin Clark, Minh-Thang Luong, Quoc~V Le, and Christopher~D Manning. 2020.
\newblock {ELECTRA}: Pre-training text encoders as discriminators rather than
  generators.
\newblock In \emph{ICLR}.

\bibitem[{Cubuk et~al.(2019)Cubuk, Zoph, Mane, Vasudevan, and
  Le}]{cubuk2019autoaugment}
Ekin~D Cubuk, Barret Zoph, Dandelion Mane, Vijay Vasudevan, and Quoc~V Le.
  2019.
\newblock {AutoAugment}: Learning augmentation strategies from data.
\newblock In \emph{CVPR}, pages 113--123.

\bibitem[{Devlin et~al.(2019)Devlin, Chang, Lee, and
  Toutanova}]{devlin2018bert}
Jacob Devlin, Ming-Wei Chang, Kenton Lee, and Kristina Toutanova. 2019.
\newblock {BERT}: Pre-training of deep bidirectional transformers for language
  understanding.
\newblock In \emph{NAACL}.

\bibitem[{Donsker and Varadhan(1975)}]{donsker1975asymptotic}
Monroe~D Donsker and SR~Srinivasa Varadhan. 1975.
\newblock Asymptotic evaluation of certain markov process expectations for
  large time, i.
\newblock \emph{Communications on Pure and Applied Mathematics}, 28(1):1--47.

\bibitem[{Edunov et~al.(2018)Edunov, Ott, Auli, and
  Grangier}]{edunov2018understanding}
Sergey Edunov, Myle Ott, Michael Auli, and David Grangier. 2018.
\newblock Understanding back-translation at scale.
\newblock \emph{arXiv preprint arXiv:1808.09381}.

\bibitem[{Erickson et~al.(2020)Erickson, Mueller, Shirkov, Zhang, Larroy, Li,
  and Smola}]{erickson2020autogluon}
Nick Erickson, Jonas Mueller, Alexander Shirkov, Hang Zhang, Pedro Larroy,
  Mu~Li, and Alexander Smola. 2020.
\newblock Autogluon-tabular: Robust and accurate automl for structured data.
\newblock \emph{arXiv preprint arXiv:2003.06505}.

\bibitem[{Fakoor et~al.(2020)Fakoor, Mueller, Erickson, Chaudhari, and
  Smola}]{fakoor2020fast}
Rasool Fakoor, Jonas~W Mueller, Nick Erickson, Pratik Chaudhari, and
  Alexander~J Smola. 2020.
\newblock Fast, accurate, and simple models for tabular data via augmented
  distillation.
\newblock In \emph{Advances in Neural Information Processing Systems},
  volume~33.

\bibitem[{Guo et~al.(2019)Guo, Mao, and Zhang}]{guo2019augmenting}
Hongyu Guo, Yongyi Mao, and Richong Zhang. 2019.
\newblock Augmenting data with mixup for sentence classification: An empirical
  study.
\newblock \emph{arXiv preprint arXiv:1905.08941}.

\bibitem[{Gupta and Agrawal(2020)}]{gupta2020compression}
Manish Gupta and Puneet Agrawal. 2020.
\newblock Compression of deep learning models for text: A survey.
\newblock \emph{arXiv preprint arXiv:2008.05221}.

\bibitem[{Hinton et~al.(2014)Hinton, Vinyals, and Dean}]{hinton2015distilling}
Geoffrey Hinton, Oriol Vinyals, and Jeff Dean. 2014.
\newblock Distilling the knowledge in a neural network.
\newblock In \emph{NIPS 2014 Deep Learning Workshop}.

\bibitem[{Hou et~al.(2020)Hou, Huang, Shang, Jiang, Chen, and
  Liu}]{hou2020dynabert}
Lu~Hou, Zhiqi Huang, Lifeng Shang, Xin Jiang, Xiao Chen, and Qun Liu. 2020.
\newblock {DynaBERT}: Dynamic bert with adaptive width and depth.
\newblock In \emph{Advances in Neural Information Processing Systems}.

\bibitem[{Hutter et~al.(2014)Hutter, Hoos, and
  Leyton-Brown}]{hutter2014efficient}
Frank Hutter, Holger Hoos, and Kevin Leyton-Brown. 2014.
\newblock An efficient approach for assessing hyperparameter importance.
\newblock In \emph{ICML}, pages 754--762. PMLR.

\bibitem[{Jiao et~al.(2020)Jiao, Yin, Shang, Jiang, Chen, Li, Wang, and
  Liu}]{jiao2019tinybert}
Xiaoqi Jiao, Yichun Yin, Lifeng Shang, Xin Jiang, Xiao Chen, Linlin Li, Fang
  Wang, and Qun Liu. 2020.
\newblock {TinyBERT}: Distilling {BERT} for natural language understanding.
\newblock In \emph{EMNLP}.

\bibitem[{Kaplan et~al.(2020)Kaplan, McCandlish, Henighan, Brown, Chess, Child,
  Gray, Radford, Wu, and Amodei}]{kaplan2020scaling}
Jared Kaplan, Sam McCandlish, Tom Henighan, Tom~B Brown, Benjamin Chess, Rewon
  Child, Scott Gray, Alec Radford, Jeffrey Wu, and Dario Amodei. 2020.
\newblock Scaling laws for neural language models.
\newblock \emph{arXiv preprint arXiv:2001.08361}.

\bibitem[{Kim et~al.(2021)Kim, Oh, Kim, Cho, and Yun}]{kim2021comparing}
Taehyeon Kim, Jaehoon Oh, NakYil Kim, Sangwook Cho, and Se-Young Yun. 2021.
\newblock Comparing kullback-leibler divergence and mean squared error loss in
  knowledge distillation.
\newblock In \emph{IJCAI}.

\bibitem[{Kong et~al.(2020)Kong, d'Autume, Ling, Yu, Dai, and
  Yogatama}]{kong2019mutual}
Lingpeng Kong, Cyprien de~Masson d'Autume, Wang Ling, Lei Yu, Zihang Dai, and
  Dani Yogatama. 2020.
\newblock A mutual information maximization perspective of language
  representation learning.
\newblock In \emph{ICLR}.

\bibitem[{Li et~al.(2020)Li, Liu, Zhao, Xu, Yang, and Jin}]{li2020bert}
Jianquan Li, Xiaokang Liu, Honghong Zhao, Ruifeng Xu, Min Yang, and Yaohong
  Jin. 2020.
\newblock {BERT}-{EMD}: Many-to-many layer mapping for bert compression with
  earth mover{'}s distance.
\newblock In \emph{EMNLP}.

\bibitem[{Liang et~al.(2021)Liang, Hao, Shen, Zhou, Chen, Chen, and
  Carin}]{liang2020mixkd}
Kevin~J Liang, Weituo Hao, Dinghan Shen, Yufan Zhou, Weizhu Chen, Changyou
  Chen, and Lawrence Carin. 2021.
\newblock {MixKD}: Towards efficient distillation of large-scale language
  models.
\newblock In \emph{ICLR}.

\bibitem[{Liu et~al.(2019)Liu, Ott, Goyal, Du, Joshi, Chen, Levy, Lewis,
  Zettlemoyer, and Stoyanov}]{liu2019roberta}
Yinhan Liu, Myle Ott, Naman Goyal, Jingfei Du, Mandar Joshi, Danqi Chen, Omer
  Levy, Mike Lewis, Luke Zettlemoyer, and Veselin Stoyanov. 2019.
\newblock {RoBERTa}: A robustly optimized bert pretraining approach.
\newblock \emph{arXiv preprint arXiv:1907.11692}.

\bibitem[{Mirzadeh et~al.(2020)Mirzadeh, Farajtabar, Li, Levine, Matsukawa, and
  Ghasemzadeh}]{mirzadeh2020improved}
Seyed~Iman Mirzadeh, Mehrdad Farajtabar, Ang Li, Nir Levine, Akihiro Matsukawa,
  and Hassan Ghasemzadeh. 2020.
\newblock Improved knowledge distillation via teacher assistant.
\newblock In \emph{AAAI}.

\bibitem[{Nguyen et~al.(2010)Nguyen, Wainwright, and
  Jordan}]{nguyen2010estimating}
XuanLong Nguyen, Martin~J Wainwright, and Michael~I Jordan. 2010.
\newblock Estimating divergence functionals and the likelihood ratio by convex
  risk minimization.
\newblock \emph{IEEE Transactions on Information Theory}, 56(11):5847--5861.

\bibitem[{Oord et~al.(2018)Oord, Li, and Vinyals}]{oord2018representation}
Aaron van~den Oord, Yazhe Li, and Oriol Vinyals. 2018.
\newblock Representation learning with contrastive predictive coding.
\newblock \emph{arXiv preprint arXiv:1807.03748}.

\bibitem[{Pennington et~al.(2014)Pennington, Socher, and
  Manning}]{pennington2014glove}
Jeffrey Pennington, Richard Socher, and Christopher~D. Manning. 2014.
\newblock \href {http://www.aclweb.org/anthology/D14-1162} {Glove: Global
  vectors for word representation}.
\newblock In \emph{EMNLP}, pages 1532--1543.

\bibitem[{Poole et~al.(2019)Poole, Ozair, Van Den~Oord, Alemi, and
  Tucker}]{poole2019variational}
Ben Poole, Sherjil Ozair, Aaron Van Den~Oord, Alex Alemi, and George Tucker.
  2019.
\newblock On variational bounds of mutual information.
\newblock In \emph{ICML}.

\bibitem[{Raffel et~al.(2019)Raffel, Shazeer, Roberts, Lee, Narang, Matena,
  Zhou, Li, and Liu}]{raffel2019exploring}
Colin Raffel, Noam Shazeer, Adam Roberts, Katherine Lee, Sharan Narang, Michael
  Matena, Yanqi Zhou, Wei Li, and Peter~J Liu. 2019.
\newblock Exploring the limits of transfer learning with a unified text-to-text
  transformer.
\newblock \emph{arXiv preprint arXiv:1910.10683}.

\bibitem[{Rajpurkar et~al.(2016)Rajpurkar, Zhang, Lopyrev, and
  Liang}]{rajpurkar2016squad}
Pranav Rajpurkar, Jian Zhang, Konstantin Lopyrev, and Percy Liang. 2016.
\newblock {SQuAD}: 100,000+ questions for machine comprehension of text.
\newblock In \emph{EMNLP}.

\bibitem[{Sanh et~al.(2019)Sanh, Debut, Chaumond, and
  Wolf}]{sanh2019distilbert}
Victor Sanh, Lysandre Debut, Julien Chaumond, and Thomas Wolf. 2019.
\newblock {DistilBERT}, a distilled version of {BERT}: smaller, faster, cheaper
  and lighter.
\newblock \emph{arXiv preprint arXiv:1910.01108}.

\bibitem[{Shi et~al.(2021)Shi, Mueller, Erickson, Li, and
  Smola}]{shi2021multimodal}
Xingjian Shi, Jonas Mueller, Nick Erickson, Mu~Li, and Alex Smola. 2021.
\newblock Multimodal automl on structured tables with text fields.
\newblock In \emph{8th ICML Workshop on Automated Machine Learning (AutoML)}.

\bibitem[{Sun et~al.(2019)Sun, Cheng, Gan, and Liu}]{sun2019patient}
Siqi Sun, Yu~Cheng, Zhe Gan, and Jingjing Liu. 2019.
\newblock Patient knowledge distillation for {BERT} model compression.
\newblock In \emph{EMNLP}.

\bibitem[{Tschannen et~al.(2020)Tschannen, Djolonga, Rubenstein, Gelly, and
  Lucic}]{tschannen2019mutual}
Michael Tschannen, Josip Djolonga, Paul~K Rubenstein, Sylvain Gelly, and Mario
  Lucic. 2020.
\newblock On mutual information maximization for representation learning.
\newblock \emph{ICLR}.

\bibitem[{Turc et~al.(2019)Turc, Chang, Lee, and Toutanova}]{turc2019well}
Iulia Turc, Ming-Wei Chang, Kenton Lee, and Kristina Toutanova. 2019.
\newblock Well-read students learn better: On the importance of pre-training
  compact models.
\newblock \emph{arXiv preprint arXiv:1908.08962}.

\bibitem[{Vaswani et~al.(2017)Vaswani, Shazeer, Parmar, Uszkoreit, Jones,
  Gomez, Kaiser, and Polosukhin}]{vaswani2017attention}
Ashish Vaswani, Noam Shazeer, Niki Parmar, Jakob Uszkoreit, Llion Jones,
  Aidan~N Gomez, {\L}ukasz Kaiser, and Illia Polosukhin. 2017.
\newblock Attention is all you need.
\newblock In \emph{Advances in neural information processing systems}, pages
  5998--6008.

\bibitem[{Wang et~al.(2019{\natexlab{a}})Wang, Pruksachatkun, Nangia, Singh,
  Michael, Hill, Levy, and Bowman}]{wang2019superglue}
Alex Wang, Yada Pruksachatkun, Nikita Nangia, Amanpreet Singh, Julian Michael,
  Felix Hill, Omer Levy, and Samuel~R Bowman. 2019{\natexlab{a}}.
\newblock Superglue: A stickier benchmark for general-purpose language
  understanding systems.
\newblock \emph{arXiv preprint arXiv:1905.00537}.

\bibitem[{Wang et~al.(2019{\natexlab{b}})Wang, Singh, Michael, Hill, Levy, and
  Bowman}]{wang2018glue}
Alex Wang, Amanpreet Singh, Julian Michael, Felix Hill, Omer Levy, and Samuel~R
  Bowman. 2019{\natexlab{b}}.
\newblock {GLUE}: A multi-task benchmark and analysis platform for natural
  language understanding.
\newblock In \emph{ICLR}.

\bibitem[{Wei and Zou(2019)}]{wei2019eda}
Jason Wei and Kai Zou. 2019.
\newblock {EDA}: Easy data augmentation techniques for boosting performance on
  text classification tasks.
\newblock In \emph{EMNLP}.

\bibitem[{Yang et~al.(2021)Yang, Martinez, Bulat, and
  Tzimiropoulos}]{yang2020knowledge}
Jing Yang, Brais Martinez, Adrian Bulat, and Georgios Tzimiropoulos. 2021.
\newblock Knowledge distillation via softmax regression representation
  learning.
\newblock In \emph{ICLR}.

\bibitem[{Zhang et~al.(2018)Zhang, Cisse, Dauphin, and
  Lopez-Paz}]{zhang2017mixup}
Hongyi Zhang, Moustapha Cisse, Yann~N Dauphin, and David Lopez-Paz. 2018.
\newblock {Mixup}: Beyond empirical risk minimization.
\newblock In \emph{ICLR}.

\end{thebibliography}
% \FloatBarrier
% \balance{}
\clearpage
% \bibliographystyle{acl_natbib}
% \clearpage

\appendix
\twocolumn[
\begin{center}
{\Large \textbf{Appendix -- \  
Distiller: A Systematic Study of Model Distillation Methods \\ \hspace*{-33mm}  in Natural Language Processing}}
\vspace*{2em}
\end{center}
]
\section{Relationship with Other Distillation Methods}

\emph{Distiller} is a generic meta-framework that encompasses various KD pipelines used in previous work. For example, Distiller with the following configurations corresponds to the KD pipeline used in each of the cited works: 
$l^{\text{pred}} =$ CE, $l^{\text{inter}}=$ MSE, $m_{i,j} = $ Skip, $a = $ CA  \cite{jiao2019tinybert}; 
$l^{\text{pred}} =$ CE, $l^{\text{inter}}=$ MSE, $m_{i,j} = $ EMD \cite{li2020bert}; 
$l^{\text{pred}} =$ CE, $a = $ Mixup  \cite{liang2020mixkd}.

\section{Proof of Theorem 3.1}
\label{sec:appdx:proof}
\noindent We denote the Mutual Information (MI) between two random variables $X$ and $Y$ as $I(X; Y)$. Based on the results on variational bounds of MI~\cite{poole2019variational}, we derived that optimizing common knowledge distillation objectives, including Mean Squared Error (MSE), L2 distance, and cosine similarity between $X$ and $Y$, can be viewed as maximizing certain lower bounds of $I(X; Y)$.

\begin{lemma}[$I_{\text{TUBA}}$].
Assume that $f(x,y)$ is an arbitrary neural network that takes $x$ and $y$ as inputs and outputs a scalar and $a(y)>0$. The lower bound of $x$ and $y$ can be estimated by:
\begin{small}
\begin{equation*}
\begin{aligned}
    I (X;Y) &\geq E_{p(x,y)}[ \log \frac{e^{f(x,y)}}{a(y)}] - E_{p(x)p(y)}[\frac{e^{f(x,y)}}{a(y)}] \\
    &\triangleq I_{\text{TUBA}}
\end{aligned}
\end{equation*}
\end{small}
\label{lemma:tuba}
\end{lemma}

% \subsection*{Lemma 1 ($I_{\text{TUBA}}$)}
% Assume that $f(x,y)$ is an arbitrary neural network that takes $x$ and $y$ as inputs and outputs a scalar and $a(y)>0$. We have:
% \begin{scriptsize}
% \begin{equation}
%     I (X;Y) \geq E_{p(x,y)}[ \log \frac{e^{f(x,y)}}{a(y)}] - E_{p(x)p(y)}[\frac{e^{f(x,y)}}{a(y)}] \triangleq I_{TUBA}
% \end{equation}
% \end{scriptsize}

\noindent\emph{Proof.}

Based on the definition of MI, we have:
\begin{equation}
    \begin{aligned}
        I(X;Y)&=E_{p(x,y)}[\log\frac{p(x|y)}{p(x)}]\\ &=E_{p(x,y)}[\log \frac{p(y|x)}{p(y)}]
    \end{aligned}
\end{equation}
  Replacing the intractable conditional distribution $p(x|y)$ with a tractable variational distribution $q(x|y)$ yields a lower bound on MI due to the non-negativity of the KL divergence:
  
\begin{scriptsize}
\begin{align}
\label{eq:tuba-proof-main}
        I(X;Y)&=E_{p(x,y)}[\log \frac{q(x|y)}{p(x)}]+E_{p(x,y)}[\log \frac{p(x|y)}{q(x|y)}] \notag\\
        &=E_{p(x,y)}[\log \frac{q(x|y)}{p(x)}]+E_{p(y)}[KL({p(x|y)}||{q(x|y)})]\\ 
        &\geq E_{p(x,y)}[\log q(x|y)]+H(X) \notag
\end{align}
\end{scriptsize}
where $H(X)$ is the entropy of $X$.\\
We choose an energy-based variational family that uses a critic $f(x,y)$ and scaled by the data density $p(x)$:
\begin{equation}
    q(x|y)=\frac{p(x)}{Z(y)}e^{f(x,y)} , Z(y)=E_{p(x)}[e^{f(x,y)}]
\end{equation}
Substituting this distribution into \eqref{eq:tuba-proof-main} gives a lower bound on MI:
\begin{center}
    $I(X;Y)\geq E_{p(x,y)}[f(x,y)]-E_{p(y)}[\log Z(y)]$
\end{center}
However, this objective is still intractable. To form a tractable bound, we can upper bound the log partition function by this inequality: $\log(x)\leq \frac{x}{a}+\log(a)-1$ for all $x,a>0$. Apply this inequality to get:

\begin{scriptsize}
\begin{equation}
    \begin{split}
        I(X;Y)&\geq E_{p(x,y)}[f(x,y)]-E_{p(y)}[\log Z(y)]\\ &\geq E_{p(x,y)}[f(x,y)]\\&-E_{p(y)}[\frac{E_{p(x)}[e^{f(x,y)}]}{a(y)}+\log a(y)-1]\\
        &= E_{p(x,y)}[f(x,y)] - E_{p(x,y)}[\log a(y)] \\
        &- E_{p(x)p(y)}[\frac{e^{f(x,y)}}{a(y)}]+1\\
        &\geq E_{p(x,y)}[\log \frac{e^{f(x,y)}}{a(y)}] - E_{p(x)p(y)}[\frac{e^{f(x,y)}}{a(y)}]
    \end{split}
\end{equation}
\end{scriptsize}
This bound holds for any $a(y)>0$

\paragraph*{Theorem 1. }Minimizing the MSE, L2, PKD loss, and maximizing the cosine similarity between two random variables $x$,$y$ can be viewed as maximizing lower bounds of $I(X;Y)$. In knowledge distillation, $x$ and $y$ are hidden states generated by student model and teacher model.\\\\
\emph{Proof.}
We prove this theorem by constructing $f(x,y)$ and $a(y)$ in Lemma \ref{lemma:tuba} for each loss function. 
\subparagraph{MSE}
$L_{\text{MSE}}(x,y) = ||x-y||_2^2$, let $f(x,y) = -||x-y||_2^2, a(y)=1$, we have:
\begin{scriptsize}
\begin{equation}
\begin{split}
 I (X;Y)&\geq E_{p(x,y)}[\log e^{-||x-y||_2^2}] - E_{p(x)p(y)}[e^{-||x-y||_2^2}] \notag \\
&\geq  E_{p(x,y)}[\log e^{-||x-y||_2^2}] - E_{p(x)p(y)}[e^0]\\ 
&= E_{p(x,y)}[\log e^{-||x-y||_2^2}] - 1\\
&= E_{p(x,y)}[-||x-y||_2^2] - 1.
\end{split}
\end{equation}
\end{scriptsize}

Thus, minimizing the MSE loss between $x$ and $y$ can be viewed as maximizing the lower bound of $I(X;Y)$.
\subparagraph{PKD Loss}
$L_{\text{PKD}}(x, y) = ||\frac{x}{||x||_2}-\frac{y}{||y||_2}||_2^2$, let $f(x,y)=-||\frac{x}{||x||_2}-\frac{y}{||y||_2}||_2^2$, we have:
\begin{scriptsize}
\begin{equation}
\begin{split}
 I (X;Y)&\geq E_{p(x,y)}[\log e^{-||\frac{x}{||x||_2}-\frac{y}{||y||_2}||_2^2}] \\
 &- E_{p(x)p(y)}[e^{-||\frac{x}{||x||_2}-\frac{y}{||y||_2}||_2^2}] \\
&\geq  E_{p(x,y)}[\log e^{-||\frac{x}{||x||_2}-\frac{y}{||y||_2}||_2^2}] - E_{p(x)p(y)}[e^0] \\ 
&= E_{p(x,y)}[\log e^{-||\frac{x}{||x||_2}-\frac{y}{||y||_2}||_2^2}] - 1\\
&= E_{p(x,y)}[-||\frac{x}{||x||_2}-\frac{y}{||y||_2}||_2^2] - 1 .
\end{split}
\end{equation}
\end{scriptsize}

Thus, minimizing the PKD loss between $x$ and $y$ can be viewed as maximizing the lower bound of $I(X;Y)$.

\subparagraph{L2 Loss}

$L_{L_2}(x,y) = ||x-y||_2$, let $f(x,y)=-||x-y||_2$, a(y)=1, we have:
\begin{scriptsize}
\begin{equation}
\begin{split}
I (X;Y)&\geq  E_{p(x,y)}[\log e^{-||x-y||_2}] - E_{p(x)p(y)}[e^{-||x-y||_2}] \notag \\
&\geq  E_{p(x,y)}[\log e^{-||x-y||_2}] - E_{p(x)p(y)}[e^0] \notag \\
&= E_{p(x,y)}[\log e^{-||x-y||_2}] - 1\\&= E_{p(x,y)}[-||x-y||_2] - 1
\end{split}
\end{equation}
\end{scriptsize}

Thus, minimizing L2 loss between $x$ and $y$ is equivalent to maximizing the lower bound of $I(X,Y)$.
\subparagraph{Cosine Similarity}
The cosine similarity between two hidden states $x$ and $y$ is calculated as $\frac{x \cdot y}{||x||\times||y||}$, let $f(x,y)=\frac{x \cdot y}{||x||\times||y||}-1, a(y)=1$
\begin{scriptsize}
\begin{equation*}
\begin{split}
I(X,Y)&\geq E_{p(x,y)}[\log e^{\frac{x \cdot y}{||x||\times||y||}-1}]-E_{p(x)p(y)}[e^{\frac{x \cdot y}{||x||\times||y||}-1}]\\
&\geq E_{p(x,y)}[\log e^{\frac{x \cdot y}{||x||\times||y||}-1}]-E_{p(x)p(y)}[e^0] \\
&= \frac{1}{e}E_{p(x,y)}[\log e^{\frac{x \cdot y}{||x||\times||y||}}]-1 \\ &= \frac{1}{e}E_{p(x,y)}[\frac{x \cdot y}{||x||\times||y||}]-1
\end{split}
\end{equation*}
\end{scriptsize}
Thus, maximizing cosine similarity between $x$ and $y$ can be viewed as maximizing mutual information between $x$ and $y$.

% \section{Mixup for Sentence Tagging}
% In Section \ref{subsec:DA}, we propose to extend mixup from classification task to sentence tagging task. To verify this, we conduct experiments on sentence tagging task SQuAD v1.1 and use mixup to generate virtual samples. For the \emph{Distiller} configurations, we fix the intermediate loss as MI-$\alpha$ $(\alpha=0.9)$, intermediate layer mapping as Skip, KD loss as CE and mixup as the only DA approach used. The teacher model is finetuned from a $\text{BERT}_\text{BASE}$ model. Three students $\text{ELECTRA}_\text{SMALL}$, $\text{TinyBERT}_4$ and $\text{TinyBERT}_6$ are initialize either from task-agnostic distillation \cite{clark2020electra} or pretrained from scratch \cite{jiao2019tinybert}. Results can be found in Table \ref{tab:mi-alpha-squad}. We can tell that mixup promotes the performance of distillation especially when there is a better teacher $\text{ELECTRA}_{\text{LARGE}}$. This observation proves that mixup gives more opportunity for the student to learn from the teacher, and also verify the effectiveness of our proposed mixup on sentence tagging task.

\begin{table*}[]
\caption{Comparison of evaluation results on GLUE test set.  We compare the distillation performance when using MLP and Transformer as critic functions in MI-$\alpha$ respectively. $\text{BERT}_{\text{BASE}}$~\text{(T)} indicates the teacher model trained by ourselves. Both of the students are initialized with $\text{TinyBERT}_4$~\cite{jiao2019tinybert} and distilled with ``Skip'' as intermediate layer mapping strategy and MI-$\alpha$ as intermediate objective functions. The difference is, $\text{TinyBERT}_4$~(MLP) is trained with a 4-layer MLP with hidden state of 512 as critic function while $\text{TinyBERT}_4$~(Transformer) uses a 2-layer Transformer with feed-forward hidden size 256. The result shows that a small Transformer architecture performs as a better critic function than an MLP in MI-$\alpha$ especially when the task is a token-level task (SQuAD v1.1).} 
\resizebox{0.95\linewidth}{!}{%
\begin{tabular}{l|lcccccccccc}
\toprule
Model & MNLI-m & MNLI-mm & QQP & QNLI & SST-2 & CoLA & MRPC & RTE & STS-B & SQuAD v1.1 & AVG \\
 & (393k) & (393k) & (364k) & (108k) & (67k) & (8.5k) & (3.5k) & (2.5k) & (5.7k) & \multicolumn{1}{c}{(108k)} &  \\ \hline
$\text{BERT}_{\text{BASE}}$~\text{(T)} & 84.5 & 83.6 & 71.7 & 90.9 & 93.4 & 49.3 & 87.0 & 67.3 & 84.7 & 88.2 & 80.0 \\ \hline
$\text{TinyBERT}_4$~(MLP) & 81.7 & \textbf{80.6} & 69.7 & \textbf{87.6} & \textbf{91.6} & 24.9 & \textbf{87.0} & \textbf{67.4} & 81.9 & 70.1 & 74.3 \\
$\text{TinyBERT}_4$~(Transformer) & \textbf{81.9} & \textbf{80.6} & \textbf{69.8} & 87.4 & 91.5 & \textbf{25.9} & \textbf{87.0} & \textbf{67.4} & \textbf{84.0} & \textbf{71.7} & \textbf{74.7} \\
\bottomrule
\end{tabular}}
\label{tab:MI-critics}
\end{table*}
\begin{table*}[!tb]
\caption{Comparison of evaluation results on GLUE dev set. We compare the distillation performance with(out) intermediate distillation. A $\text{BERT}_{\text{BASE}}$ model is used as the teacher and $\text{TinyBERT}_4$ is the student. $\text{TinyBERT}_4$~(KD) represents using a vanilla knowledge distillation~(student only learns from the outputs of teacher) and ``+intermediate distillation'' represents using vanilla KD and intermediate distillation.} 
\resizebox{0.95\linewidth}{!}{%
\begin{tabular}{l|lcccccccccc}
\toprule
Model & MNLI-m & MNLI-mm & QQP & QNLI & SST-2 & CoLA & MRPC & RTE & STS-B & SQuAD v1.1 & AVG \\
 & (393k) & (393k) & (364k) & (108k) & (67k) & (8.5k) & (3.5k) & (2.5k) & (5.7k) & \multicolumn{1}{c}{(108k)} &  \\
 & acc & acc & f1/acc & acc & acc & mcc & f1/acc & acc & spearman/pearson & f1/em &  \\ \hline
$\text{BERT}_{\text{BASE}}$~\text{(T)} & 84.1 & 84.7 & 88.0/91.1 & 91.7 & 93.0 & 55.0 & 89.6/85.0 & 65.0 & 88.4/88.6 & 88.2/80.9 & 83.8 \\ \hline
$\text{TinyBERT}_4$~(KD) & 80.1 & 80.3 & 86.4/89.7 & 85.8 & 89.1 & 16.1 & \textbf{89.6}/\textbf{85.3} & \textbf{66.8} & \textbf{88.4}/\textbf{88.5} & 77.3/66.8 & 77.9 \\
+intermediate distillation & \textbf{80.7} & \textbf{81.3} & \textbf{87.0}/\textbf{90.2} & \textbf{86.8} & \textbf{90.0} & \textbf{21.3} & 89.3/84.8 & 65.3 & 88.2/88.4 & \textbf{79.4}/\textbf{69.4} & \textbf{78.7} \\
\bottomrule
\end{tabular}}
\label{tab:intermediate distillation}
\end{table*}
\section{Architecture of Teacher and Student Networks}
\label{sec:app:netarch}
In Table \ref{tab:mi-alpha}, we use two baseline models $\text{BERT-PKD}_4$ and $\text{BERT-EMD}_4$. As described in the original paper \cite{sun2019patient}, we initialize $\text{BERT-PKD}_4$ with the first 4 layers of parameters from pretrained $\text{BERT}_\text{BASE}$. $\text{BERT-EMD}_4$ is initialized from $\text{TinyBERT}_4$ so they have the same architecture.
We list the detailed configurations of the teacher and student architectures investigated in our study in Table \ref{tab:teacher_student_architecture}.
\begin{figure}[!tb]
  \centering
    \includegraphics[scale=0.4]{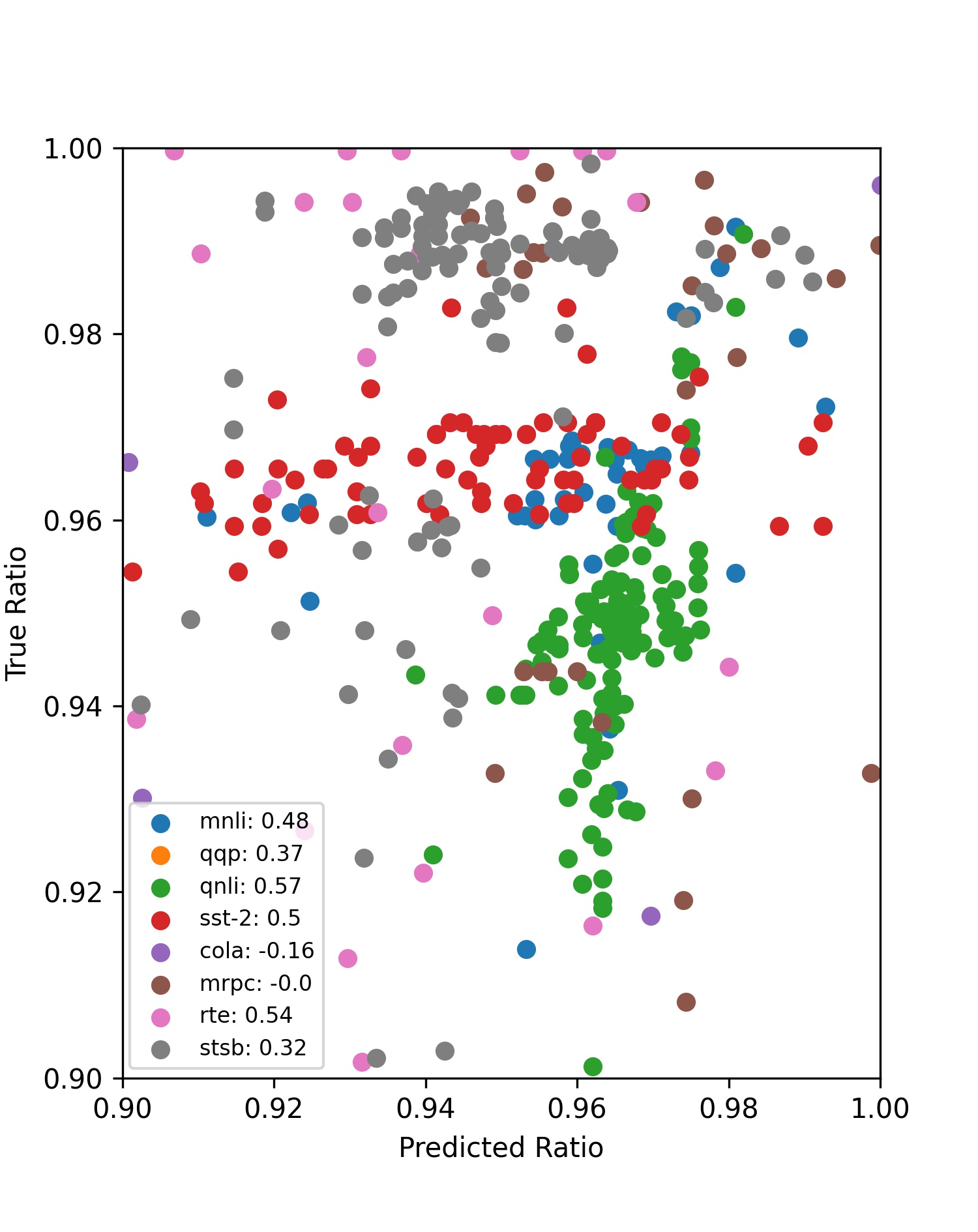}
  \caption{Evaluating held-out AutoDistiller predictions on GLUE via leave-one-out estimates. We use dataset-level cross-validation that holds out each GLUE dataset from AutoDistiller training. For each held-out dataset, the legend lists Spearman's rank correlation between the predicted vs.\ actual distillation ratio across different KD pipelines. The average Spearman's rank correlation value across the 8 datasets is 0.33.
  }
%   \JM{Number in legend needs to be defined. I presume these are some sort of rank correlation, if not, why not report rank correlation somewhere?}
  \label{fig:leaveoneout}
\end{figure}
\begin{figure}[b]
%    \begin{minipage}{6cm}
    \centering
    \vspace{-1em}
    \includegraphics[width=1.0\columnwidth]{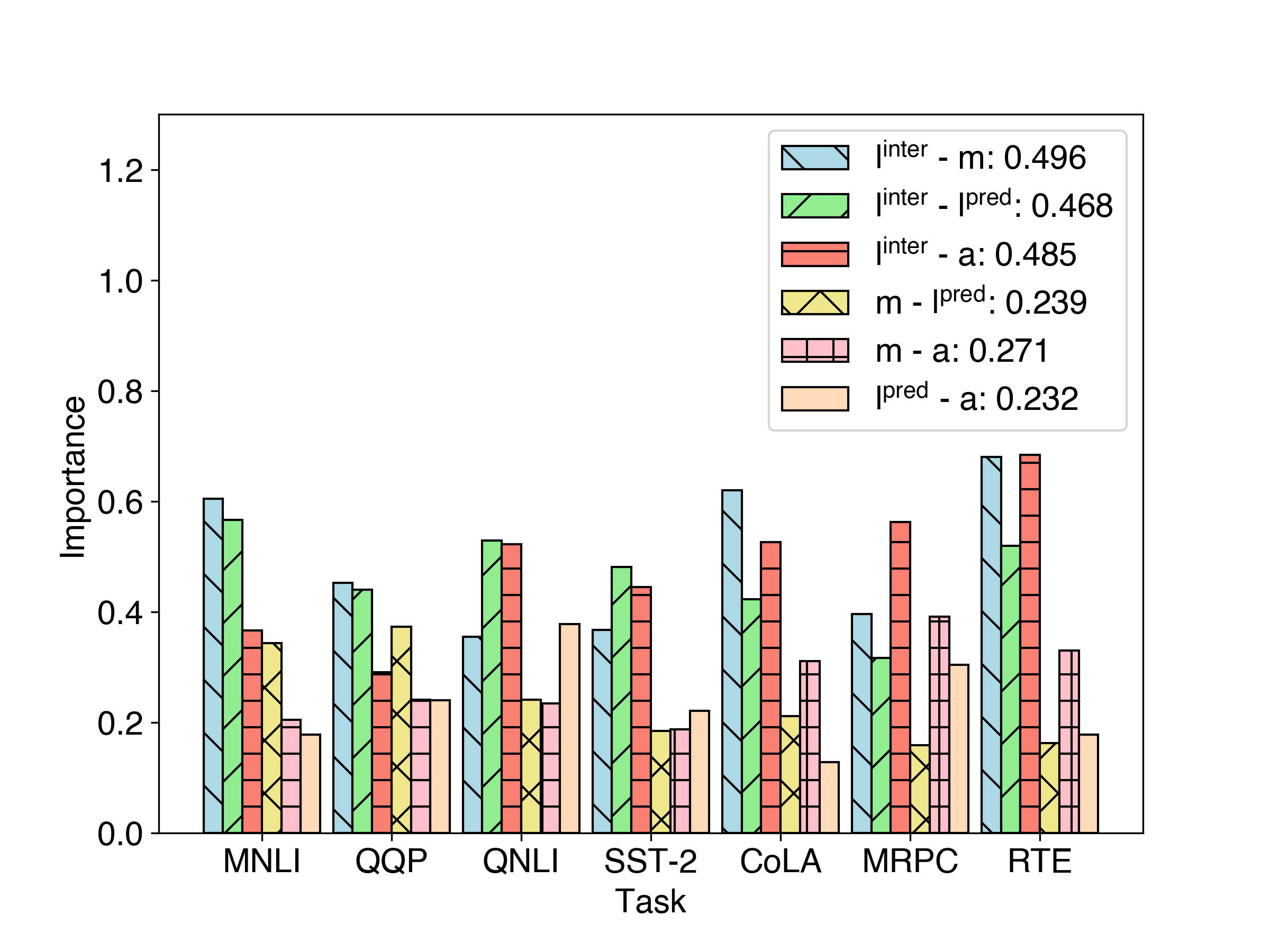}
%    \end{minipage}
    \caption{Importance of interactions between hyper-parameters.}
    \label{fig:fanova-interactions}
\end{figure}
\begin{table}[h]
\centering
\caption{Network architectures of the teacher and student models used in the paper.}
\resizebox{0.95\columnwidth}{!}{%
\begin{tabular}{lccccc}
\toprule
Model & \#Params &  $h_{\text{units}}$ & $n_{\text{layers}}$ & $h_{\text{mid}}$ & $n_{\text{heads}}$ \\
\hline
\multicolumn{6}{c}{Teacher architectures} \\
$\text{RoBERTa}_{\text{LARGE}}$ & 335M & 1024 & 24 & 4096 & 16 \\
$\text{ELECTRA}_{\text{LARGE}}$ & 335M & 1024 & 24 & 4096 & 16 \\
$\text{BERT}_{\text{BASE}}$ & 110M & 768 & 12 & 3072 & 12 \\ \hline
\multicolumn{6}{c}{Student architectures} \\
$\text{TinyBERT}_{\text{6}}$ & 67M & 768 & 6 & 3072 & 12 \\
$\text{BERT-PKD}_4$ & 52M & 768 & 4 & 3072 & 12 \\ 
$\text{BERT}_{\text{MEDIUM}}$ & 41M & 512 & 8 & 2048 & 8 \\
$\text{BERT}_{\text{SMALL}}$ & 29M & 512 & 4 & 2048 & 8 \\
$\text{TinyBERT}_{\text{4}}$ & 14M & 312 & 4 & 1200 & 12 \\
$\text{ELECTRA}_{\text{SMALL}}$ & 14M & 256 & 12 & 1024 & 4 \\
$\text{BERT}_{\text{MINI}}$ & 11M & 256 & 4 & 1024 & 4 \\
$\text{BERT}_{\text{TINY}}$ & 4M & 128 & 2 & 512 & 2 \\
\bottomrule
\end{tabular}
}
\label{tab:teacher_student_architecture}
\end{table}
\section{Benefits of Intermediate Distillation}
Table \ref{tab:intermediate distillation} shows the experimental results of distillation with(out) intermediate distillation across different tasks/datasets. All the experiments are trained 20 epochs and the intermediate distillation constitutes of ``Skip'' as intermediate mapping strategy and CE as intermediate distillation object.

\section{Dataset Embedding}
\label{sec:app:task_embedding}
As features for the AutoDistiller model, we extract features from datasets and represent them as a fixed dimension embedding. Details are shown in Table \ref{tab:task_embedding}.

\begin{table*}[tb!]
\centering
\caption{We extract features from downstream datasets so that every task can be represented as a fixed-dimension embedding. The extracted embedding can be fed into \emph{AutoDistiller} as dense features. In this table, we describe how the embedding is acquired. }
\resizebox{0.95\linewidth}{!}{
\begin{tabular}{l|l}
\toprule
Feature & Description \\ \hline
Context Embedding & \begin{tabular}[c]{@{}l@{}} Every document can be represented as a weighted average of the GloVe vectors,\\ where the weights are defined by the TF-IDF scheme. \\
Each down-stream dataset is viewed as a ``document'' in the TF-IDF scheme. \\ Precisely, the embedding of a dataset $s$ is $v_s = \frac{1}{|s|}\text{IDF}_w v_w$, \\
where $|s|$ denotes the number of words in the dataset, \\
and $\text{IDF}_w:=\log \frac{1+N}{1+N_w}$ is the inverse document frequency of word $w$. \\
$N$ is the total number of datasets, and $N_w$ denotes the number of datasets containing $w$. \\ Intuitively, this feature represents the content of datasets.\end{tabular} \\ \hline
Task Embedding & \begin{tabular}[c]{@{}l@{}}For the dataset, we collect their literal descriptions, usually one or two sentences.\\ Then aggregate GloVe vectors of every words in these sentences and get a description embedding.\\ This feature is used to represent what is the NLP task and how the data is formatted.\end{tabular} \\ \hline
Baseline Score & \begin{tabular}[c]{@{}l@{}}We use a lite Bi-LSTM model as the baseline model and finetune it\\ on the down-stream dataset. \\ This feature aims to measure the difficulty of each task by measuring \\ how well a simple architecture can perform on the specific dataset.\end{tabular} \\ \hline
Teacher Score & \begin{tabular}[c]{@{}l@{}} The fine-tuned teacher score on the dataset. Comparing teacher score to aforementioned \\baseline score tells how much boost can a complex model has on this dataset.\end{tabular}\\ \hline
Number of Examples & How many training samples in the dataset. \\ \bottomrule
\end{tabular}}
\label{tab:task_embedding}
\end{table*}

\section{How much does larger/better teacher help?}
Table \ref{tab:teacher} shows the performance of different students distilled from teachers of different sizes and pretraining schemes. From the results, we found that although the teacher $\text{ELECTRA}_{\text{LARGE}}$ has the best performance on average score, most of the students of $\text{ELECTRA}_{\text{LARGE}}$ performs worse than students of $\text{RoBERTa}_{\text{LARGE}}$.~$\text{ELECTRA}_{\text{SMALL}}$ is the only student that performs the best with $\text{ELECTRA}_{\text{LARGE}}$ as teacher, that may be attributed to $\text{ELECTRA}_{\text{SMALL}}$ and $\text{ELECTRA}_{\text{LARGE}}$ are pretrained on the same pretraining task, so they have a similar knowledge representation scheme. And also, for datasets~(MNLI, QQP, QNLI and SST-2) which have 
abundant amount of data, students of $\text{BERT}_{\text{BASE}}$ perform better. 
\begin{table*}[tb!]
\centering
\caption{Performance comparison with different teacher and student models. We abbreviate three teacher models $\text{BERT}_{\text{BASE}}$,$~\text{RoBERTa}_{\text{LARGE}}$ and $\text{ELECTRA}_{\text{LARGE}}$ as B, R, E. Results are evaluated on GLUE dev set and best results are in-bold.} 
\resizebox{0.95\textwidth}{!}{%
\begin{tabular}{lccccccccccccccc}
\hline
Model & \#Params & Teacher & MNLI-m & MNLI-mm & \multicolumn{2}{c}{QQP} & QNLI & SST-2 & CoLA & \multicolumn{2}{c}{MRPC} & RTE & \multicolumn{2}{c}{STS-B} & AVG \\
 &  &  & acc & acc & f1 & acc & acc & acc & mcc & f1 & acc & acc & spearman & pearson &  \\ \hline
$\text{BERT}_{\text{BASE}}~\text{(T)}$ & 110M &  & 84.1 & 84.7 & 88.0 & 91.1 & 91.7 & 93.0 & 55.0 & 89.6 & 85.0 & 65.0 & 88.4 & 88.6 & 83.7 \\
$\text{RoBERTa}_{\text{LARGE}}~\text{(T)}$ & 335M &  & 90.2 & 90.1 & 89.6 & 92.1 & 94.7 & 96.3 & 64.6 & 91.3 & 88.0 & 78.7 & \textbf{91.7} & \textbf{91.8} & 88.3 \\
$\text{ELECTRA}_{\text{LARGE}}~\text{(T)}$ & 335M &  & \textbf{90.5} & \textbf{90.4} & \textbf{90.3} & \textbf{92.8} & \textbf{95.1} & \textbf{96.6} & \textbf{67.4} & \textbf{91.7} & \textbf{88.5} & \textbf{84.5} & 88.7 & 88.9 & \textbf{88.8} \\ \hline
$\text{BERT}_{\text{BASE}}$ & 110M & $\text{R}$ & \textbf{84.5} & \textbf{84.6} & 88.6 & 91.5 & \textbf{91.7} & \textbf{93.2} & 59.3 & 91.6 & 88.0 & 66.4 & 89.0 & 89.4 & 84.8 \\
 &  & $\text{E}$ & 84.4 & \textbf{84.6} & \textbf{88.8} & \textbf{91.7} & 91.6 & 92.8 & \textbf{59.5} & \textbf{91.9} & \textbf{88.7} & \textbf{69.3} & \textbf{89.1} & \textbf{89.6} & \textbf{85.2} \\ \hline
$\text{TinyBERT}_{\text{6}}$ & 67M & $\text{B}$ & \textbf{83.9} & \textbf{84.0} & \textbf{88.1} & \textbf{91.2} & \textbf{91.3} & 91.6 & \textbf{50.5} & 90.3 & 86.5 & 75.5 & 89.4 & 89.4 & \textbf{84.3} \\
 &  & $\text{R}$ & 83.5 & 83.5 & 88.0 & \textbf{91.2} & 90.8 & \textbf{92.2} & 48.0 & \textbf{91.9} & \textbf{88.7} & 72.6 & \textbf{89.9} & \textbf{90.0} & 84.2 \\
 &  & $\text{E}$ & 83.0 & 83.0 & 87.8 & 91.0 & 90.6 & 91.3 & 48.6 & 91.6 & 88.5 & \textbf{76.2} & 89.1 & 89.3 & 84.2 \\ \hline
$\text{BERT}_{\text{MEDIUM}}$ & 41M & $\text{B}$ & \textbf{82.6} & \textbf{83.0} & \textbf{87.9} & \textbf{91.0} & \textbf{90.0} & 90.8 & 48.3 & 88.9 & 84.1 & \textbf{65.0} & \textbf{88.2} & 88.4 & \textbf{82.4} \\
 &  & $\text{R}$ & 80.9 & 81.4 & 87.6 & 90.8 & 89.0 & \textbf{91.4} & 50.5 & 88.9 & 84.6 & 64.3 & \textbf{88.2} & \textbf{88.6} & 82.2 \\
 &  & $\text{E}$ & 81.0 & 81.3 & 87.5 & 90.7 & 89.0 & 90.9 & \textbf{51.0} & \textbf{89.5} & \textbf{85.3} & 64.3 & 88.0 & 88.2 & 82.2 \\ \hline
$\text{BERT}_{\text{SMALL}}$ & 29M & $\text{B}$ & \textbf{81.0} & \textbf{81.0} & \textbf{87.4} & \textbf{90.6} & \textbf{87.3} & \textbf{90.5} & \textbf{43.1} & 87.8 & 82.4 & 63.5 & 87.0 & 87.2 & \textbf{80.7} \\
 &  & $\text{R}$ & 78.7 & 78.6 & 87.0 & 90.4 & 87.0 & 88.6 & 41.2 & \textbf{89.1} & \textbf{84.1} & \textbf{64.3} & \textbf{87.1} & \textbf{87.3} & 80.3 \\
 &  & $\text{E}$ & 78.6 & 78.8 & 87.2 & 90.5 & 87.0 & 89.3 & 43.0 & 88.7 & \textbf{84.1} & 63.9 & 86.8 & 87.1 & 80.4 \\ \hline
$\text{TinyBERT}_{\text{4}}$ & 14M & $\text{B}$ & \textbf{81.1} & \textbf{81.6} & \textbf{87.2} & \textbf{90.4} & \textbf{87.4} & \textbf{90.6} & 12.3 & 89.4 & 85.0 & 66.4 & 87.7 & 87.8 & 78.9 \\
 &  & $\text{R}$ & 80.0 & 80.7 & 86.5 & 90.0 & 86.0 & 89.4 & \textbf{24.6} & 90.4 & 86.5 & 67.9 & \textbf{88.0} & \textbf{88.1} & \textbf{79.8} \\
 &  & $\text{E}$ & 80.0 & 80.2 & 86.2 & 89.6 & 85.9 & 88.9 & 21.8 & \textbf{90.9} & \textbf{86.8} & \textbf{68.6} & 87.6 & 87.6 & 79.5 \\ \hline
$\text{ELECTRA}_{\text{SMALL}}$ & 14M & $\text{B}$ & \textbf{82.7} & \textbf{83.8} & 87.8 & 90.9 & \textbf{89.7} & 91.3 & \textbf{60.6} & 91.3 & 87.7 & 60.6 & 87.4 & 87.5 & 83.5 \\
 &  & $\text{R}$ & 82.3 & 83.2 & 88.1 & 91.2 & 89.5 & 90.6 & 58.6 & 91.3 & 87.5 & 67.5 & \textbf{87.6} & \textbf{87.8} & 83.8 \\
 &  & $\text{E}$ & 82.0 & 82.7 & \textbf{88.5} & \textbf{91.5} & 89.3 & \textbf{91.4} & \textbf{60.6} & \textbf{92.3} & \textbf{89.0} & \textbf{69.7} & 86.6 & 86.7 & \textbf{84.2} \\ \hline
$\text{BERT}_{\text{MINI}}$ & 11M & $\text{B}$ & \textbf{78.5} & \textbf{79.7} & \textbf{86.6} & \textbf{90.0} & \textbf{84.9} & \textbf{87.8} & 20.1 & \textbf{87.0} & \textbf{81.6} & 61.0 & \textbf{86.2} & 86.1 & 77.5 \\
 &  & $\text{R}$ & 76.6 & 77.3 & 86.0 & \textbf{90.0} & 84.6 & 85.9 & 32.2 & 86.6 & 81.1 & \textbf{65.3} & \textbf{86.2} & \textbf{86.3} & \textbf{78.2} \\
 &  & $\text{E}$ & 76.3 & 77.1 & 85.9 & 89.5 & 84.2 & 85.9 & \textbf{33.8} & \textbf{87.0} & \textbf{81.6} & 65.0 & 85.7 & 85.5 & 78.1 \\ \hline
$\text{BERT}_{\text{TINY}}$ & 4M & $\text{B}$ & \textbf{72.8} & \textbf{73.4} & \textbf{83.5} & \textbf{87.2} & \textbf{81.3} & \textbf{83.9} & 0.0 & 84.5 & 75.7 & 58.5 & 81.4 & \textbf{79.9} & 71.8 \\
 &  & $\text{R}$ & 71.5 & 72.0 & 82.9 & 86.7 & 80.2 & 83.1 & 6.2 & 84.9 & 76.2 & 60.3 & \textbf{81.9} & \textbf{79.9} & \textbf{72.1} \\
 &  & $\text{E}$ & 71.2 & 71.8 & 82.9 & 87.0 & 80.0 & 82.8 & \textbf{6.7} & \textbf{85.2} & \textbf{77.0} & \textbf{62.8} & 78.4 & 77.3 & 71.9 \\ \hline
\end{tabular}}
\label{tab:teacher}
\end{table*}
\section{Computing Details}
\label{sec:app:com-config}
All the experiments are performed on a single machine with 4 NVIDIA T4 GPUs. For hyper-parameters in Equation~\ref{eq:objective}, our experiments suggest that setting $\beta_1$ and $\beta_2$ to $1$ produces the best overall performance so we fix their values to $1$ in subsequent results. $\gamma_1$ and $\gamma_2$ are set to $0.5$ when DA is used (otherwise $\beta_2$, $\gamma_1$ and $\gamma_2$ are set to $0$). For controlled experiments, unless specified explicitly, we fix ${l^\text{inter}}$ as MI-$\alpha$ $(\alpha=0.9)$ and the layer mapping as ``Skip''. Critic functions in MI-$\alpha$ are powered by a two-layer Transformer($h_\text{mid}=256$, $n_\text{heads}=8$) and the comparison of using Transformer and MLP to estimate critic functions is described in Table \ref{tab:MI-critics}. To reduce the hyper-parameter search space, we fix the batch size as 16 and the learning rate as 5e-5 for all experiments. We used automated mix precision for training. Maximum sequence length is set to 128 for sentence-pair tasks in GLUE, 64 for single sentence tasks and 320 for SQuAD. Most of the experiments are trained for 20 epochs except 50 epochs for the challenging task CoLA and 30 epochs for SQuAD. 

\section{Distiller Search Space}
\label{sec:app:searchspace}
Here we recap the full search space considered for each stage of the KD pipeline in Distiller:
\begin{itemize}
    \item $l^{\text{inter}}$ $\in$ \{MSE, L2, Cos, PKD, MI-$\alpha$ ($\alpha$ = 0.1, 0.5, or 0.9)\}
    \item $l^{\text{pred}}$ $\in$ \{MSE, CE\}
    \item \{$m_{i,j}$\} $\in$ \{Skip, Last, EMD\}
    \item Augmentation policy $a$ is one or combinations of elementary augmentation operations in \{CA, RA, BT, Mixup\}.
\end{itemize}
%$l^{\text{inter}}$ $\in$ \{MSE, L2, Cos, PKD, MI-$\alpha$ ($\alpha$ = 0.1, 0.5, or 0.9)\}, $l^{\text{pred}}$ $\in$ \{MSE, CE\}, \{$m_{i,j}$\} $\in$ \{Skip, Last, EMD\}, augmentation policy $a$ is one or combinations of elementary augmentation operations in \{CA, RA, BT, Mixup\}.

\section{Top Configurations in Distiller}
\label{sec:app:top-config}
Here we list the top 5 configurations from the Distiller search space that performed best on  each dataset in Table~\ref{tab:top5}.
\begin{table*}[tb!]
\centering
\caption{Top 5 configurations to distill a $\text{BERT}_{\text{BASE}}$ teacher to a $\text{TinyBERT}_4$ student on every dataset. To reduce the search space, we only compare configurations that don't use data augmentation. As Hyper-parameter $\alpha$ is only valid for MI-$\alpha$, the value of $\alpha$ is set to N for other intermediate loss in the table.}
\resizebox{0.95\linewidth}{!}{%
\begin{tabular}{lclccll}
\toprule
Task & Intermediate Loss  & $\alpha$ & Layer Mapping Strategy & KD Loss & \#Example & Score \\
\hline \hline
MNLI & MI-$\alpha$ & 0.9 & EMD & MSE & 393000 & 81.7 \\
MNLI & MI-$\alpha$ & 0.1 & EMD & MSE & 393000 & 81.7 \\
MNLI & MI-$\alpha$ & 0.5 & Skip & MSE & 393000 & 81.7 \\
MNLI & MSE & N & EMD & MSE & 393000 & 81.6 \\
MNLI & MSE & N & Skip & MSE & 393000 & 81.6 \\ \hline
QQP & MI-$\alpha$ & 0.9 & EMD & MSE & 364000 & 90.2 \\
QQP & MI-$\alpha$ & 0.1 & EMD & MSE & 364000 & 90.2 \\
QQP & CE & N & Skip & MSE & 364000 & 90.2 \\
QQP & MI-$\alpha$ & 0.1 & EMD & MSE & 364000 & 90.1 \\
QQP & MI-$\alpha$ & 0.1 & Skip & MSE & 364000 & 90.1 \\ \hline
QNLI & CE & N & Last & MSE & 105000 & 87.4 \\
QNLI & MI-$\alpha$ & 0.5 & Skip & MSE & 105000 & 87.4 \\
QNLI & MI-$\alpha$ & 0.5 & Last & CE & 105000 & 87.3 \\
QNLI & MI-$\alpha$ & 0.9 & Skip & CE & 105000 & 87.2 \\
QNLI & MI-$\alpha$ & 0.1 & Skip & CE & 105000 & 87.1 \\
\hline
SST-2 & Cos & N & Last & MSE & 67000 & 90.6 \\
SST-2 & MSE & N & Skip & CE & 67000 & 90.5 \\
SST-2 & MI-$\alpha$ & 0.1 & Last & CE & 67000 & 90.3 \\
SST-2 & CE & N & Skip & MSE & 67000 & 90.3 \\
SST-2 & MI-$\alpha$ & 0.9 & Skip & CE & 67000 & 90.3 \\ \hline
CoLA & MI-$\alpha$ & 0.1 & EMD & MSE & 8500 & 22.3 \\
CoLA & MI-$\alpha$ & 0.5 & EMD & MSE & 8500 & 21.6 \\
CoLA & MI-$\alpha$ & 0.5 & EMD & CE & 8500 & 21.1 \\
CoLA & MI-$\alpha$ & 0.1 & Last & MSE & 8500 & 21.1 \\
CoLA & MI-$\alpha$ & 0.5 & Skip & MSE & 8500 & 21.0 \\ \hline
MRPC & MI-$\alpha$ & 0.1 & EMD & CE & 3700 & 90.3 \\
MRPC & MI-$\alpha$ & 0.5 & EMD & CE & 3700 & 90.2 \\
MRPC & CE & N & Skip & CE & 3700 & 89.9 \\
MRPC & MI-$\alpha$ & 0.9 & EMD & CE & 3700 & 89.9 \\
MRPC & CE & N & Last & MSE & 3700 & 89.7 \\ \hline
RTE & MI-$\alpha$ & 0.1 & Skip & CE & 2500 & 70.8 \\
RTE & MI-$\alpha$ & 0.9 & Skip & CE & 2500 & 70.4 \\
RTE & MI-$\alpha$ & 0.5 & Skip & CE & 2500 & 70.0 \\
RTE & MI-$\alpha$ & 0.1 & Last & CE & 2500 & 70.0 \\
RTE & MI-$\alpha$ & 0.5 & Last & CE & 2500 & 69.3 \\\hline
STS-B & MI-$\alpha$ & 0.9 & Skip & MSE & 7000 & 88.0 \\
STS-B & MI-$\alpha$ & 0.5 & Skip & MSE & 7000 & 88.0 \\
STS-B & MI-$\alpha$ & 0.9 & EMD & MSE & 7000 & 87.9 \\
STS-B & MI-$\alpha$ & 0.1 & Last & MSE & 7000 & 87.9 \\
STS-B & PKD & N & Skip & MSE & 7000 & 87.9 \\
\hline
SQuAD v1.1 & MSE & N & Skip & MSE & 130000 & 72.6 \\
SQuAD v1.1 & CE & N & Skip & MSE & 130000 & 72.4 \\
SQuAD v1.1 & MSE & N & EMD & MSE & 130000 & 72.3 \\
SQuAD v1.1 & MI-$\alpha$ & 0.9 & Skip & CE & 130000 & 71.9 \\
SQuAD v1.1 & MI-$\alpha$ & 0.9 & Skip & CE & 130000 & 71.7 \\
 \bottomrule
\end{tabular}}
\label{tab:top5}
\end{table*} 
% \todo{Organize this table and draw a figure of top five intermediate loss}

\end{document}